\documentclass[lettersize,journal]{IEEEtran}
\usepackage{amsmath,amsfonts}
\usepackage{algorithmic}
\usepackage{array}
\usepackage[caption=false,font=normalsize,labelfont=sf,textfont=sf]{subfig}
\usepackage{textcomp}
\usepackage{stfloats}
\usepackage{subcaption}
\usepackage{url}
\usepackage{verbatim}
\usepackage{graphicx}
\usepackage{subcaption}
\usepackage{algorithm, algorithmic}
\usepackage{cite}
\usepackage{balance}
\usepackage{gensymb}
\usepackage{amssymb}

\newtheorem{ass}{\textbf{Assumption}}
\newtheorem{pro}{\textbf{Proposition}}

\newtheorem{dnt}{\textbf{Definition}}
\newtheorem{rem}{\textbf{Remark}}

\newtheorem{prb}{\textbf{Problem}}

\def\BibTeX{{\rm B\kern-.05em{\sc i\kern-.025em b}\kern-.08em
    T\kern-.1667em\lower.7ex\hbox{E}\kern-.125emX}}
\usepackage{balance}
\begin{document}
\title{Hierarchical Motion Planning and Control under Unknown Nonlinear Dynamics via Predicted Reachability}
\author{Zhiquan Zhang$^{\dagger}$ and Melkior Ornik$^{\dagger}$
\thanks{This research was supported in part by NASA under grant number 80NSSC22M0070 and by the Air Force Office of Scientific Research under grant number FA9550-23-1-0131.}
\thanks{$^{\dagger}$Zhiquan Zhang and Melkior Ornik are with the University of Illinois Urbana-Champaign, IL, 61801 USA. Emails:
        {\tt\small \{zz121, mornik\}@illinois.edu}}
        }


\maketitle

\begin{abstract}
Autonomous motion planning under unknown nonlinear dynamics requires learning system properties while navigating toward a target. In this work, we develop a hierarchical planning-control framework that enables online motion synthesis with limited prior system knowledge. The state space is partitioned into polytopes and approximates the unknown nonlinear system using a piecewise-affine (PWA) model. The local affine models are identified once the agent enters the corresponding polytopes. To reduce computational complexity, we introduce a non-uniform adaptive state space partition strategy that refines the partition only in task-relevant regions. The resulting PWA system is abstracted into a directed weighted graph, whose edge existence is incrementally verified using reach control theory and predictive reachability conditions. Certified edges are weighted using provable time-to-reach bounds, while uncertain edges are assigned information-theoretic weights to guide exploration. The graph is updated online as new data becomes available, and high-level planning is performed by graph search, while low-level affine feedback controllers are synthesized to execute the plan. Furthermore, the conditions of classical reach control theory are often difficult to satisfy in underactuated settings. We therefore introduce relaxed reachability conditions to extend the framework to such systems. Simulations demonstrate effective exploration-exploitation trade-offs with formal reachability guarantees.
\end{abstract}


\section{Introduction}
Robotic motion planning and control under prior unknown environments pose fundamental challenges in a wide range of applications, such as mobile ground and aerial navigation~\cite{shim2005autonomous, mustafa2019towards, zhou2022swarm, zhang2024efp}, autonomous exploration in planetary and extraterrestrial missions~\cite{10783051}, indoor robotic exploration in complex built environments~\cite{8737705}. In these settings, robots are often required to make decisions with limited or uncertain information about the environment, while simultaneously coping with constrained sensing capabilities, potentially hazardous and poorly structured dynamics, and restricted onboard computational resources~\cite{dixit2024step, ahmadi2020risk, ono2015chance}. 

A large amount of literature has investigated autonomous exploration in unknown environments while satisfying high-level task objectives. Early approaches typically decouple exploration and exploitation. For example,~\cite{6697120} proposed a frontier-based exploration strategy combined with incremental motion planning to identify feasible trajectories that satisfy a given task specification, treating information gathering and task execution as separate phases. Subsequent work sought to unify these two aspects by reasoning over probabilistic representations of the environment. In particular,~\cite{9304481} performed planning on a probabilistic graph that incorporates prior environmental knowledge, thereby softening the classical separation between exploration and exploitation. The work of~\cite{9197522} introduced a map-predictive motion planning framework that explicitly accounts for uncertainty in the environment by predicting future map evolution and incorporating it into the planning process. 

However, existing methods predominantly model uncertainty at a high, symbolic level, such as unknown occupancy or state labels, while largely assuming known or well-behaved low-level system dynamics. As a result, uncertainty arising from unknown or partially known robot dynamics, and the associated challenges in control synthesis and reachability guarantees under such uncertainty, remain insufficiently addressed.

This work considers the problem of driving a robot to a desired target while its underlying dynamics, dictated by both its mechanical properties and possibly uncertain environmental features, are not known, except a bound on their rate of change. The difficulty in solving this problem arises from two aspects. On the one hand, achieving the target requires determining whether the robot can transition between regions of the state space, i.e., whether certain states or boundaries are reachable. However, performing reachability analysis with constrained control inputs is naturally nontrivial when system dynamics are unknown~\cite{shafa2022reachability}. On the other hand, the robot must determine an effective decision-making strategy that jointly address online system identification and controller synthesis, balancing exploration of the unknown dynamics with exploitation of the acquired knowledge during mission execution. 

In addressing control problems with unknown dynamics, the control community has focused on data-driven methodologies for online system identification and controller synthesis. Reinforcement learning-based approaches, for instance, have been used to approximate optimal control policies that steer a system toward a desired equilibrium, accompanied by Lyapunov-based stability guarantees~\cite{6042339}. In parallel, adaptive control techniques~\cite{landau2011adaptive, bar1988simple} aim to synthesize stabilizing controllers while estimating uncertain system parameters in real time. 

Beyond purely deterministic formulations, another line of research incorporates probabilistic representations to capture uncertainty in the dynamics~\cite{9610137, 6385724, 6024480}. For example,~\cite{6024480} proposed a receding-horizon control framework that jointly performs prediction, estimation, and planning under chance constraints. The work of~\cite{9610137} also explored navigation in unknown environments. It presented a framework of online incremental mapping and a multi-layered planning scheme that addresses robotic motion constraints and probabilistic environmental constraints to guarantee safe navigation. However, these approaches primarily focus on parameter learning within structured system models and presuppose the reachability of the target state. Furthermore, many existing control approaches focus primarily on asymptotic convergence to an equilibrium state, without explicitly accounting for the structure and constraints of the resulting trajectories. This limits their applicability to complex robotic tasks that require interactive behaviors and restricts potential extensions to sequential task specifications. 

A recent work~\cite{11245807} proposed a framework of motion planning and control for systems that combine reach specifications with unknown low-level dynamics. It developed a unified hierarchical planning-control framework that operates online. The proposed approach represents the state space using a collection of polytopic regions and locally approximates the unknown nonlinear dynamics by affine models, yielding a piecewise-affine abstraction. Transitions between neighboring regions are captured by a directed graph whose edges encode facet reachability, enabling high-level planning over the abstracted state space while low-level controllers are synthesized to execute the selected transitions. 

While this framework demonstrates certain effectiveness, it suffers from several limitations, such as high computational complexity and difficulties in handling underactuated systems. To address these issues, this paper presents several significant advances:
\begin{enumerate}
    \item in order to address the exponential computational growth induced by uniform state space discretization as the system dimension or operating domain increases, we propose an adaptive non-uniform state space partitioning strategy that selectively refines regions relevant to the task, significantly reducing the cost of repeated reachability analysis;
    \item the construction of the graph constructed by facet reachability is enhanced by an edge-weighting mechanism for uncertain transitions, where weights are assigned based on the expected information gain quantified via entropy, enabling an explicit and interpretable exploration–exploitation trade-off during planning;
    \item the framework is extended to underactuated mobile robots by introducing suitable relaxations of reachability conditions for a certain set of dynamics, allowing controller synthesis under actuation constraints while retaining conservative guarantees, and validated through corresponding simulation studies.
\end{enumerate}
Together, these contributions substantially improve the scalability, expressiveness, and applicability of reachability-based motion planning under unknown dynamics, enabling efficient online planning and control synthesis for both fully actuated and underactuated robotic systems.

The remainder of this paper is organized as follows. Section~\ref{sec:prelim} introduces necessary mathematical preliminaries and notations. Section~\ref{sec:pf} presents formal formulation of the problem. Section~\ref{sec:approach} presents our hybrid motion planning framework. It integrates path planning and control synthesis and incorporates an adaptive non-uniform state space partitioning strategy. Section~\ref{sec:case} presents representative case studies: (i) a fully actuated ground mobile robot operating over unknown terrain and (ii) an underactuated robotic system that requires suitable theoretical relaxations within the proposed framework. Section~\ref{sec:conclusion} concludes the paper.

\section{Preliminaries and Notations}\label{sec:prelim}
We first outlines the essential preliminaries and notational conventions, including definitions related to transition systems, directed weighted graphs, and polytopes, along with the mathematical notations and terminology used in this paper.
\subsection{Transition Systems and Directed Weighted Graphs}
A deterministic transition system~\cite{baier2008principles} is defined as a six-tuple $\mathcal{TS} = (S, Act, \rightarrow, I, AP, L)$, where $S$ is a non-empty set of \textit{states} $s \in S$ and $Act$ is a non-empty set of \textit{actions} $a \in Act$. The transition relation $\rightarrow \subseteq S \times Act \times S$ specifies the possible state transitions, where $(s, a, s') \in \rightarrow$ denotes that the system can move from state $s$ to $s'$ upon executing action $a$. The set $AP$ contains \textit{atomic propositions} describing properties of the states, and the \textit{labeling function} $L:S \rightarrow 2^{AP}$ assigns to each state $s \in S$ the subset $L(s) \subseteq AP$ of propositions that hold true in that state.

A finite \textit{path} (or \textit{run}) from an initial state $s_0 \in I$ is a sequence
$\pi = s_0 \stackrel{a_0}{\longrightarrow} s_1 \stackrel{a_1}{\longrightarrow} s_2 \stackrel{a_2}{\longrightarrow} \ldots s_{N_\pi}$, with $(s_i, a_i, s_{i+1}) \in \rightarrow$ for all $i < N_{\pi}$. A state $s' \in S$ is \textit{reachable} from $s_0$ if $s'$ appears in a path originating from $s_0$.

A transition system is represented by a \textit{directed weighted graph} $\mathcal{G}=(V, E, w, AP, L)$. The set of \textit{vertices} $V$ corresponds to the set of states $S$ in the transition system. The set of \textit{edges} $E$ consists of all pairs of states $(s, s')$ such that there exists an action $a \in Act$ with $(s, a, s') \in \rightarrow$. Each transition is associated with a weight function $c(s,s'): E\rightarrow \mathbb{R}$ that assigns a real-valued weight to each edge. For a path $\pi$ in the transition system, its corresponding weight in the graph is given by
\begin{equation}
    w(\pi) = \sum_i c(s_i, s_{i+1}).
\end{equation}
The labeling function $L$ and the set of atomic propositions $AP$ are inherited directly from the corresponding transition system.

\subsection{Polytopes}
Polytopes can be characterized by the following two equivalent definitions \cite{ziegler2012lectures}.

\begin{dnt}
    A set $\mathcal{P} \subseteq \mathbb{R}^n$ is called a \textit{full-dimensional polytope} is there exists a finite set of points $\{v_1, v_2, \ldots, v_m\} \subset \mathbb{R}^n$ $(m\ge n+1)$ such that 
    \begin{equation}
        \mathcal{P} = {\rm conv}\{v_i\}_{i=1\ldots m} = \left\{\sum_{i=1}^m \lambda_i v_i : \sum_{i =1}^{m} \lambda_i = 1, \lambda_i\ge 0 \right\}.
    \end{equation}
\end{dnt}
That is, a full-dimensional polytope is the convex hull of finitely many points in $\mathbb{R}^n$.
\begin{dnt}
    A set $\mathcal{P}\subseteq \mathbb{R}^n$ is also a full-dimensional polytope if and only if there exists a matrix $A\in \mathbb{R}^{k\times n}$ and a vector $b \in \mathbb{R}^k$ $(k\ge n+1)$ such that
    \begin{equation}
        \mathcal{P} = \{x \in \mathbb{R}^n : Ax \leq b\},
    \end{equation}
    and $\mathcal{P}$ is bounded.
\end{dnt}
In this paper, we only consider \textit{convex} polytopes. A hyperplane $H \subseteq \mathbb{R}^n$ is said to be a \textit{supporting hyperplane} of $\mathcal{P}$ if $\mathcal{P}$ is entirely contained in one of the closed half-spaces determined by $H$ and $\mathcal{P} \cap H \neq \varnothing$. A subset $F \subseteq \mathcal{P}$ is called a \textit{face} of $\mathcal{P}$ if there exists a supporting hyperplane $H$ of $\mathcal{P}$ such that $F = P \cap H$. A \textit{facet} is a face of $\mathcal{P}$ with dimension $n-1$, and a \textit{vertex} is a face of dimension $0$.
\subsection{Notations and Terminology}
For any differentiable function $f(x)$, we denote its gradient with respect to $x$ by $\nabla_x f(x) = \partial f(x)/\partial x$. For any vector $v$, $\|v\|$ represents the Euclidean norm. For any matrix $M$, $\|M\|$ denotes its norm, defined as $\|M\| = \sup_{x\neq 0}(\|Mx\|/\|x\|)$, where $x$ is any nonzero vector.

For clarity in the subsequent sections, we define the notion of ``\textit{facet reachability}'' from a reach control perspective~\cite{broucke2014reach, habets2004control}. \textit{Affine dynamics} refer to systems of the form $\dot x = Ax + Bu +c$, where $A$ and $B$ are matrices and $c$ is a vector. A \textit{affine state-feedback control law} refers to the control law of the form $u = Kx +d$, where the control input depends affinely on the state. A facet $F$ of a polytope $\mathcal{P}$ is said to be reachable under affine dynamics if, for any initial state $x_0 \in \mathcal{P}$, there exists a finite time $T_0>0$ and an affine state-feedback control law such that:
\begin{enumerate}
    \item the state trajectory remains entirely within $\mathcal{P}$ for all $t \in [0, T_0)$, i.e. $x(t)\in \mathcal{P}$;
    \item the trajectory reaches the facet $F$ of $\mathcal{P}$ at time $T_0$, i.e., $x(T_0) \in F$;
    \item the velocity at $T_0$ points outward along the normal direction of the facet, i.e., $n^\top \dot{x}(T_0)>0$, where $n$ is the outward normal vector of $F$.
\end{enumerate}
Since reach control under continuous feedback remains an open theoretical problem \cite{ornik2015topological}, in this work we restrict our focus to the construction of affine feedback controllers, which provide a tractable and implementable framework for analyzing facet reachability.

We also clarify the notion of fully-actuation and under-actuation. Suppose that the dynamics of a robot system are represented as $\dot x = F(x) + G(x)u$, where $G(x) \in \mathbb{R}^{n \times m}$. The system is said to be \textit{fully-actuated} if $\mathrm{rank}(G(x)) = n$ and \textit{under-actuated} if $\mathrm{rank}(G(x)) < n$.

\section{Problem Formulation}\label{sec:pf}
We consider an agent evolving under an unknown continuous-time nonlinear control-affine system described by
\begin{equation}\label{eq:ds}
    \dot{x} = f(x)+g(x)u,
\end{equation}
where $x \in \mathbb{R}^n$ denotes the system state and $u \in \mathbb{R}^m$ is the control input. The functions $f(x):\mathbb{R}^n \rightarrow \mathbb{R}^n$ and $g(x):\mathbb{R}^n \rightarrow \mathbb{R}^{n \times m}$ represent the drift vector field and input matrix, respectively. We make the following assumptions to formally state the problem.
\begin{ass}\label{ass:control_constraints}
    \textbf{(State space and control constraints)} The system state is constrained within a polytope $P_s \subseteq \mathbb{R}^n$, and the control input is constrained within a polytope $P_u\subseteq \mathbb{R}^m$. We assume that all trajectories of interest remain inside $P_s$ and all admissible control inputs remain inside $P_u$. Behavior outside these sets is not considered in this work.
\end{ass}
\begin{ass}\label{ass:Lip}
    \textbf{(Lipschitz continuity)} The functions $\nabla_x f(x)$ and $g(x)$ in \eqref{eq:ds} are Lipschitz continuous on $P_s$. Specifically, there exist constants $\mathcal{L}_{df}$ and $\mathcal{L}_g$ such that
    \begin{equation}
    \begin{aligned}
        \|\nabla_x f(x_1) - \nabla_x f(x_2)\| &\leq \mathcal{L}_{df} \|x_1-x_2\|\\
        \|g(x_1)-g(x_2)\| &\leq \mathcal{L}_g \|x_1 - x_2\|
    \end{aligned}
    \end{equation}
    for any $x_1, x_2 \in P_s$
\end{ass}
Assumption~\ref{ass:control_constraints} reflects a usual situation in practical applications, where system states and control inputs are subject to physical limitations such as actuator saturation, kinematic bounds, energy constraints, etc ~\cite{10610232, 9304031, TEEL1992165}. Assumption~\ref{ass:Lip} is automatically satisfied for any functions $f(x)$ and $g(x)$ that are $\mathcal{C}^2$, provided that $P_s$ is compact \cite{bartle2000introduction}. This condition ensures that, for any sufficiently smooth control input $u$, the system~\eqref{eq:ds} results in existence and uniqueness of trajectories~\cite{hsieh2012basic}.

In this paper, we consider the following problem:
\begin{prb}
    Consider an agent evolving in a nonlinear system described by \eqref{eq:ds}, where the functions $f(x)$ and $g(x)$ are unknown, but their Lipschitz constants $\mathcal{L}_{df}$ and $\mathcal{L}_g$ are known. Given an initial state $x(0) = x_0$, where $x_0 \in P_s$, and a target state $x^* \in P_s$, design a control input $u$ such that the agent reaches $x^*$ in finite time.
\end{prb}

This problem is inherently challenging due to the unknown system dynamics, which make it difficult to synthesize feasible control inputs. Addressing it requires rigorous reachability analysis while effectively balancing system identification (exploration) and motion planning and control synthesis (exploitation). While one could consider employing robust control techniques, classical robust control designs typically presuppose a known nominal model with structured uncertainty bounds and do not directly address the problem of verifying reachability under unknown dynamics during online system model learning.

To address these challenges, our approach consists of two main steps: (1) approximating the unknown nonlinear dynamics with a piecewise-affine (PWA) system, and (2) formulating a version of Problem 1 tailored to the approximated dynamics, referred to as Problem 2.

We begin by partitioning the state space $P_s$ into a finite collection of disjoint polytopes $\{P_l\}_{l\in L}$, where $L$ is a index set. Within each polytope $P_l$, the system dynamics are locally approximated by an affine model, resulting in an overall continuous-time PWA representation. The facets of these polytopes correspond to the boundaries where the system transitions between different affine regimes.

The linearization of nonlinear dynamics \eqref{eq:ds} around the equilibrium point $(x_e, u_e=0)$, where $x_e$ is an arbitrary state in $P_l$, is given by
\begin{equation}\label{eq:PWA}
\begin{aligned}
f(x) + g(x)u \approx & f(x_e) + g(x_e)u_e + \nabla_x f(x_e)(x - x_e) \\
& + \nabla_x g(x_e)(x - x_e)u_e + g(x_e)(u - u_e) \\
= & \nabla_x f(x_e)x + g(x_e)u + f(x_e) - \nabla_x f(x_e)x_e \\
= & \bar A_l x + \bar B_l u + \bar c_l,
\end{aligned}
\end{equation}
where $\bar A_l = \nabla_x f(x_e) \in \mathbb{R}^{n \times n}$, $\bar B_l = g(x_e) \in \mathbb{R}^{n \times m}$, and $\bar c_l = f(x_e) - \nabla_x f(x_e)x_e \in \mathbb{R}^n$. Naturally, if the partition of $P_s$ into polytopes $\{P_l\}$ is sufficiently fine, the maximal deviation between the true dynamics and their affine approximation can be made arbitrarily small~\cite{khalil2002nonlinear}.

We associate the partitioned state space with a directed weighted graph $\mathcal{G}_s = (V_s, E_s, w_s)$ to capture the topological structure induced by the polytope decomposition. The vertex set $V_s$ corresponds to the index set $L$ of the polytopes, such that each vertex represents a distinct region $P_l \subseteq P_s$. A directed edge $(l, l') \in E_s$ is included if there exists a facet $F_l$ of $P_l$ that is reachable from any initial state $x_{l0}\in P_l$ and this facet $F_l$ is contained in a facet of the adjacent polytope $P_{l'}$, thereby ensuring feasible transition between $P_{l}$ and $P_{l'}$. Let $P_i$ and $P_*$ denote the polytopes containing the initial state $x_0$ and the target state $x^*$, respectively. Due to the unknown nature of the system dynamics, the reachability relations between polytopes are uncertain, meaning that the existence of edges $E_s$ is initially unknown to the agent.

By representing the partitioned state space as the graph $\mathcal{G}_s$, we reformulate the original control problem as a joint path-planning and control-synthesis task over this abstracted structure. This motivates the following problem statement.

\begin{prb}
    Consider an agent evolving in an unknown nonlinear system that is locally approximated by the piecewise-affine model in \eqref{eq:PWA}, where the state space is partitioned into disjoint polytopes. Design an affine feedback controller that drives the agent from the initial state $x_i$ within polytope $P_i$ to the target polytope $P_*$. Upon entering $P_*$, design controller that drives the agent to $x^*$.
\end{prb}

This formulation establishes a structured framework that integrates graph-based motion planning with local affine control synthesis, enabling systematic reasoning about how to drive the system from $x_0$ to $x^*$ through the piecewise-affine approximations of the dynamics.
\section{Proposed Approach}\label{sec:approach}
The proposed method follows a hierarchical design that integrates two interdependent layers. At the lower level, the framework performs system identification and reachability analysis over individual polytopes to characterize local dynamics and reachability prediction. The upper level operates on a graph abstraction. It is built from the partitioned state space and local affine approximations of the dynamics. Edge existence is determined via predictive reachability analysis. To enhance computational scalability, we develop an adaptive state space partitioning algorithm that enables non-uniform resolution across the domain. Within this hierarchy, the identified affine models and local reachability results continuously refine the graph's edge information, whereas the high-level planner determines which regions of the state space the agent should explore next.

Subsection \ref{subsec:sysid} describes the procedure of local affine system identification. Subsection \ref{subsec:reachability} revisits the facet reachability conditions and affine feedback controller synthesis procedure originally proposed in \cite{habets2004control}, and extends them to predict facet reachability of polytopes with unknown dynamics by leveraging available prior knowledge of the system. Subsection \ref{subsec:partitioning} introduces a non-uniform state space partitioning algorithm that adaptively divides the state space according to the significance of different regions, thereby reducing the computational complexity of reachability prediction. Subsection \ref{subsec:graph} describes the construction of the graph abstraction, in which heuristic edge weights are assigned to edges with both certain and uncertain existence. This graph construction enables the agent to adaptively balance exploration and exploitation when navigating through the state space. Finally, subsection \ref{subsec:framework} presents an integrated motion planning and control framework that selects the next polytope to visit, while synthesizing the corresponding low-level affine feedback controllers to execute the planned motion.

\subsection{Local Affine System Identification}\label{subsec:sysid}
As the agent enters a new polytope, the local affine system identification follows a standard small-signal excitation and regression procedure, as described in Algorithm~\ref{alg:sysid}. The key idea is to repeatedly excite the system with small control perturbations over very short time intervals and record the corresponding state derivatives. The collected input-output data is then used in a linear regression step to estimate the parameters of the local affine model. By choosing sufficiently small excitation amplitudes and a small sampling period $T$, the identification error can be made arbitrarily small, yielding an accurate approximation of the underlying system dynamics~\cite{ornik2019control}.

\begin{algorithm}[h]
\caption{AffineSystemIdentification}
\begin{algorithmic}[1]\label{alg:sysid}
\REQUIRE Initial state $x_0 \in \mathbb{R}^n$, sampling period $T>0$
\STATE \textbf{Initialize:}
\[
X_{\text{data}} \gets \varnothing, \quad 
\dot X_{\text{data}} \gets \varnothing, \quad 
x \gets x_0
\]

\STATE Select a set of small control inputs $\{u^{(0)}, u^{(1)}, \dots, u^{(m)}\}$ 
that are affinely independent

\FOR{each $u^{(k)}$ in the control inputs set}
    \STATE Apply control $u^{(k)}$ over time interval $[t, t+T]$
    \STATE Measure next state:
    \[
    x_{\text{new}} = x(t+T)
    \]
    \STATE Approximate state derivative:
    \[
    \dot x \approx \frac{x_{\rm{new}} - x}{T}
    \]
    \STATE Append $[x^\top, (u^{(k)})^\top, 1]^\top$ to $X_{\text{data}}$
    \STATE Append $\dot x$ to $\dot X_{\rm{data}}$
\ENDFOR

\STATE Estimate affine parameters using least-squares:
\[
[A \; B \; c]
=
\dot X_{\text{data}}
X_{\text{data}}^\top
\left(
X_{\text{data}} X_{\text{data}}^\top
\right)^{-1}
\]

\RETURN $A$, $B$, $c$
\end{algorithmic}
\end{algorithm}
\begin{figure*}[t]
  \centering
  \includegraphics[width=\textwidth]{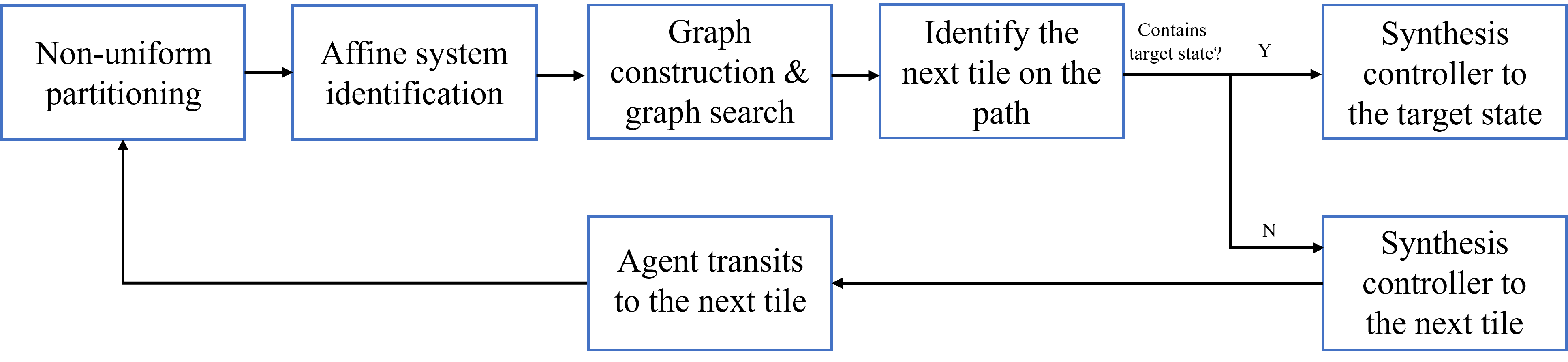} 
  \caption{A pipeline of the proposed motion planning and control framework.}
  \label{fig:pipeline}
\end{figure*}
\subsection{Facet Reachability and Controller Synthesis}\label{subsec:reachability}
To perform predictive reachability analysis using the approximated local affine system model, we firstly revisit the results from reach control theory, then introduce the predictive reachability conditions under unknown dynamics.
\subsubsection{Revisit of Facet Reachability and Feedback Controller Synthesis}
Assume that at time $t_0$ the agent's state lies within the polytope $P_l$, i.e., $x(t_0)\in P_l$. Let vertices of $P_l$ be denoted by $v_{l1}, \ldots, v_{lM}$, and its facets by $F_{l1}, \ldots, F_{lK}$, each associated with an outward-pointing normal vector $n_{l1}, \ldots, n_{lK}$. For each facet $F_{li}$, we define $V_{li}\subseteq \{1, \ldots, M\}$ as the index set of its vertices. Similarly, for each vertex $v_{lj}$, we define $W_{lj}\subseteq \{1, \ldots, K\}$ as the index set of facets that contain $v_{lj}$. We assume that the corresponding local affine dynamics $(\bar A_l, \bar B_l, \bar c_l)$ have been identified as described in Subsection \ref{subsec:sysid}. To characterize the agent's transition from $P_l$ to an adjacent polytope, we consider, without loss of generality, the reachability of facet $F_{l1}$, whose outward normal vector is $n_{l1}$. If the facet is reachable, affine state-feedback controller can be synthesized to drive the agent towards the facet and into the adjacent polytope, which indicates feasible transition between the two regions.

Suppose that there exist control inputs on the vertices $u_{l1}, \ldots, u_{lM} \in P_u$ such that the following condition hold:
\begin{subequations}\label{eq:facet-conditions}
\begin{align}
&\forall j \in V_{l1}:\nonumber\\
&\ \ \ \ \ \ \ \ \ n_{l1}^\top (\bar A_l v_{lj} + \bar B_l u_{lj} + \bar c_l) > 0;\label{eq:facet-conditions-a}\\
&\ \ \ \ \ \ \ \ \ n_{li}^\top (\bar A_l v_{lj} + \bar B_l u_{lj} + \bar c_l) \le 0,\ \forall i \in W_{lj}\setminus\{1\}; \label{eq:facet-conditions-b}\\
&\forall j \in \{1,\ldots,M\}\setminus V_{l1}:\nonumber\\
&\ \ \ \ \ \ \ \ \ n_{li}^\top (\bar A_l v_{lj} + \bar B_l u_{lj} + \bar c_l) \le 0, \forall i \in W_{lj}; \label{eq:facet-conditions-c}\\
&\ \ \ \ \ \ \ \ \ n_{l1}^\top (\bar A_l v_{lj} + \bar B_l u_{lj} + \bar c_l) > 0.\label{eq:facet-conditions-d}
\end{align}
\end{subequations}

If the above conditions are satisfied, an affine state-feedback controller can be synthesized to drive the system toward the facet $F_{l1}$ within finite time~\cite{habets2004control}. 

The controller synthesis procedure, originally proposed in~\cite{habets2004control}, is summarized as follows. We first determine a set of feasible control inputs $u_{l1}, u_{l2}, \ldots, u_{lM}$ that satisfies the above-mentioned reachability constraints \eqref{eq:facet-conditions-a}-\eqref{eq:facet-conditions-d}. Next, the polytope $P_l$ is partitioned into a collection of non-overlapping simplices whose union is $P_l$ (i.e., a triangulation of $P_l$)., From this decomposition, a simplex $\bar P_l \subset P_l$ is selected whose vertices $\bar v_{l1}, \bar v_{l2}, \ldots, \bar v_{l(n+1)}$ are each assigned an associated control input $\bar u_{l1}, \bar u_{l2}, \ldots, \bar u_{l(n+1)}$. Based on these vertex-input pairs, an affine feedback control law in simplex $\bar P_l$ is constructed as
\begin{equation}
    u(x) = F_l x + g_l,
\end{equation}
where coefficients $F_l$ and $g_l$ are obtained by solving 
\begin{equation}\label{eq:controller}
\begin{bmatrix}
F_l \mid g_l
\end{bmatrix}
\begin{bmatrix}
\bar v_{l1} & \cdots & \bar v_{l(n+1)} \\
1 & \cdots & 1
\end{bmatrix}
=
\begin{bmatrix}
\bar u_{l1} & \cdots & \bar u_{l(n+1)}
\end{bmatrix}.
\end{equation}
In essence, the affine feedback control law is a state-dependent linear combination of the feasible control inputs associated with the vertices. The resulting controller is defined locally on each simplex and therefore differs across the triangulation. When the agent's state reaches a facet of the current simplex, the control law switches to the controller associated with the adjacent simplex. Determining whether a feasible set of control inputs $\{u_{l1}, u_{l2}, \ldots, u_{lM}\}$ exists reduces to finding a feasible point within a polytope defined by the intersection of the hyperplanes specified in constraints \eqref{eq:facet-conditions-a}-\eqref{eq:facet-conditions-d} and the control constraint polytope $P_u$. This feasibility problem can be efficiently solved as it can be formulated as a linear program.

The above facet reachability analysis applies to polytopes with known affine dynamics. We next extend this analysis to develop predictive reachability conditions for polytopes whose dynamics are unknown.
\subsubsection{Predictive Reachability Criterion with Unknown Dynamics}
To extend facet reachability analysis beyond the identified polytope $P_l$, we develop a predictive reachability criterion that determines whether the facet reachability of a polytope $P_{l'}$, whose local affine dynamics are not yet known, can be guaranteed. To perform this analysis, we use the Lipschitz continuity assumption (Assumption~\ref{ass:Lip}). Since reachability depends on the feasibility of affine inequalities parameterized by $(\bar A_l, \bar B_l, \bar c_l)$, the bounded deviation between the unknown dynamics of $P_{l'}$ and those of $P_l$, derived from Assumption~\ref{ass:Lip}, can be exploited to construct conservative predictions of feasibility. Intuitively, this approach propagates knowledge from explored regions to neighboring ones while ensuring that uncertainty in the dynamics does not lead to overoptimistic reachability claims.

For simplicity of the derivation below, we need to firstly categorize the constraints in \eqref{eq:facet-conditions} into two classes: strict inequalities and non-strict inequalities. Let the unknown affine model of $P_{l'}$ be $(\bar A_{l'}, \bar B_{l'}, \bar c_{l'})$, and denote its vertices and facets by $\{v_{l'1}, \ldots, v_{l'M}\}$ and $\{F_{l'1}, \ldots, F_{l'K}\}$, respectively, with outward normals $\{n_{l'1}, \ldots, n_{l'K}\}$. We define the index sets $I_+ = \{i:n_{l'i}^\top (\bar A_{l'}v_{l'j} + \bar B_{l'}u_{l'j} + \bar c_{l'}) > 0\}$, $I_- = \{i:n_{l'i}^\top (\bar A_{l'}v_{l'j} + \bar B_{l'}u_{l'j} + \bar c_{l'}) \leq 0\}$. These sets are determined by the selection of the exit facet $\bar F_{l'}$, i.e., the facet through which the system attempts to leave $P_{l'}$. This selection fixes the associated $V_{l'i}$ and $W_{l'j}$. The reachability condition can therefore be written compactly as 
\begin{equation}\label{eq: simple condition}
    \left\{
    \begin{aligned}
    n_{l'i}^\top (\bar A_{l'} v_{l'j} + \bar B_{l'} u_{l'j} + \bar c_{l'}) &> 0, i\in I_+\\
    n_{l'i}^\top (\bar A_{l'} v_{l'j} + \bar B_{l'} u_{l'j} + \bar c_{l'}) &\le 0, i\in I_-\\
    u_{l'j} &\in P_u.
    \end{aligned}
    \right.
\end{equation}

When $(\bar A_{l'}, \bar B_{l'}, \bar c_{l'})$ are unknown, we rely on the Lipschitz bounds between $P_l$ and $P_{l'}$ to build a robust version of \eqref{eq: simple condition}. The idea is to shift the known affine parameters $(\bar A_l, \bar B_l, \bar c_l)$ within their feasible deviation limits $(\varepsilon_A, \varepsilon_B, \varepsilon_c)$, which quantify the maximal norm differences between the identified dynamics and the unknown local model under the Lipschitz assumption. This yields the most conservative feasible region that could still guarantee reachability. The following proposition formalizes upper bounds on the deviations between affine approximation at two different operating points.
\begin{pro}\label{pro:dynamics_bounds}
Suppose Assumption~\ref{ass:Lip} holds. Consider two operating points $x_1,x_2\in P_s$, and let $(\bar A_1,\bar B_1, \bar c_1)$ and $(\bar A_2,\bar B_2, \bar c_2)$ denote the corresponding affine approximations obtained at $x_1$ and $x_2$. Then the deviations between the affine parameters satisfy
\begin{equation}
    \|\bar A_2 - \bar A_1\| \leq \varepsilon_A,\ \|\bar B_2 - \bar B_1\| \leq \varepsilon_B,\ \|\bar c_2 - \bar c_1\| \leq \varepsilon_c.
\end{equation}
where
\begin{equation}
    \begin{aligned}
        \varepsilon_A &= L_{df}\|x_2 - x_1\|,\\
        \varepsilon_B &= L_g\|x_2 - x_1\|,\\
        \varepsilon_c &= \frac{1}{2}L_{df}\|x_2-x_1\|^2 + L_{df}\|x_2-x_1\|\|x_2\|.
    \end{aligned}
\end{equation}
\end{pro}
This proposition is proved in Appendix~A.


Using the dynamics deviation upper bounds introduced above, we propose the following predictive facet reachability condition.
\begin{pro}\label{prop:1}
    Consider the inequality set:
    \begin{equation}\label{eq:narrow-ineq-new}
    \scalebox{0.93}{$\left\{
    \begin{aligned}
    \bar B_{l'}^-  &u_{l'j} + n_{l'i}^\top (\bar A_{l} v_{l'j}  + \bar c_{l}) > \|n_{l'i}\|(\varepsilon_A\|v_{l'j}\| + \varepsilon_c), i\in I_+\\
     \bar B_{l'}^+ &u_{l'j} + n_{l'i}^\top (\bar A_{l} v_{l'j}  + \bar c_{l})  \leq- \|n_{l'i}\|(\varepsilon_A\|v_{l'j}\| + \varepsilon_c), i\in I_-\\
    &u_{l'j} \in P_u,
    \end{aligned}
        \right.$}
    \end{equation}
    where
        \begin{equation}
    \begin{aligned}
        \bar B^+_{l'} &= n_{l'i}^\top \bar B_l+ \left[\textbf{sgn}(u_{l'j}^1)\varepsilon_B\|n_{l'i}\|\ \cdots \textbf{sgn}(u_{l'j}^m)\varepsilon_B\|n_{l'i}\|\right ],\\
        \bar B^-_{l'} &= n_{l'i}^\top \bar B_l-\left[\textbf{sgn}(u_{l'j}^1)\varepsilon_B\|n_{l'i}\|\ \cdots\ \textbf{sgn}(u_{l'j}^m)\varepsilon_B\|n_{l'i}\|\right ],\\
    \end{aligned}
\end{equation}
    where $\textbf{sgn}(\cdot)$ is a sign function that equals to $1$ if its argument is positive, $-1$ is its argument is negative and $0$ if its argument is zero. If this inequality set admits a feasible solution, then \eqref{eq: simple condition} is feasible. As a result, the exit facet $\bar F_{l'}$ of $P_{l'}$ is reachable.
\end{pro}

A detailed proof of proposition~\ref{prop:1} is provided in Appendix~B.

\begin{rem}\label{rem:1}
The inequality set above is not affine in $u_{l'j}$ since its structure depends on the sign pattern of each component of the control input $u_{l'j} = [u_{l'j}^1,\ldots,u_{l'j}^m]^\top$. Nevertheless, the problem can be reformulated as a finite collection of affine feasibility problems by enumerating all possible sign configurations of $u_{l'j}$, resulting in $2^m$ affine inequality sets. The original feasibility problem~\eqref{eq:narrow-ineq-new} is satisfied if at least one of these $2^m$ affine inequality sets admits a feasible solution.
\end{rem}
\begin{rem}
When $\varepsilon_A = \varepsilon_B = \varepsilon_c = 0$, this formulation exactly reduces to \eqref{eq: simple condition}, as the dynamics of $P_l$ and $P_{l'}$ are identical. For non-zero deviation bounds, the robust set \eqref{eq:narrow-ineq-new} accounts for the worst-case discrepancies and therefore provides a sufficient condition for reachability under unknown affine models.
\end{rem}

To provide a counterpart to our earlier analysis, we now establish a sufficient condition for the infeasibility of~\eqref{eq: simple condition}. This condition is obtained by perturbing the identified affine model $(\bar A_l, \bar B_l, \bar c_l)$ in the opposite direction to that used for the feasibility analysis, thereby generating a feasible region of $u_{l'j}$ that is maximally expanded relative to the nominal case associated with $(\bar A_l, \bar B_l, \bar c_l)$.

\begin{pro}\label{prop:2}
    Consider the inequality set:
    \begin{equation}\label{eq:expand-ineq-new}
    \scalebox{0.93}{$\left\{
    \begin{aligned}
    \bar B_{l'}^+  &u_{l'j} + n_{l'i}^\top (\bar A_{l} v_{l'j}  + \bar c_{l}) > -\|n_{l'i}\|(\varepsilon_A\|v_{l'j}\| + \varepsilon_c), i\in I_+\\
     \bar B_{l'}^- &u_{l'j} + n_{l'i}^\top (\bar A_{l} v_{l'j}  + \bar c_{l})  \leq\|n_{l'i}\|(\varepsilon_A\|v_{l'j}\| + \varepsilon_c), i\in I_-\\
    &u_{l'j} \in P_u,
    \end{aligned}
        \right.$}
    \end{equation}
    where $\bar B_{l'}^+$ and $\bar B^-_{l'}$ are defined the same as Proposition~\ref{prop:1}. If this inequality set is infeasible, then \eqref{eq: simple condition} is guaranteed to be infeasible. Consequently, the exit facet $\bar F_{l'}$ of $P_{l'}$ is not reachable.
\end{pro}
\begin{rem}
Similar to Remark~\ref{rem:1}, the feasibility problem \eqref{eq:expand-ineq-new} can be reformulated as $2^m$ affine feasibility problems. If all such problems are infeasible, the infeasibility of problem \eqref{eq: simple condition} is guaranteed.
\end{rem}

\begin{figure}
  \centering
  \includegraphics[width=0.49\textwidth]{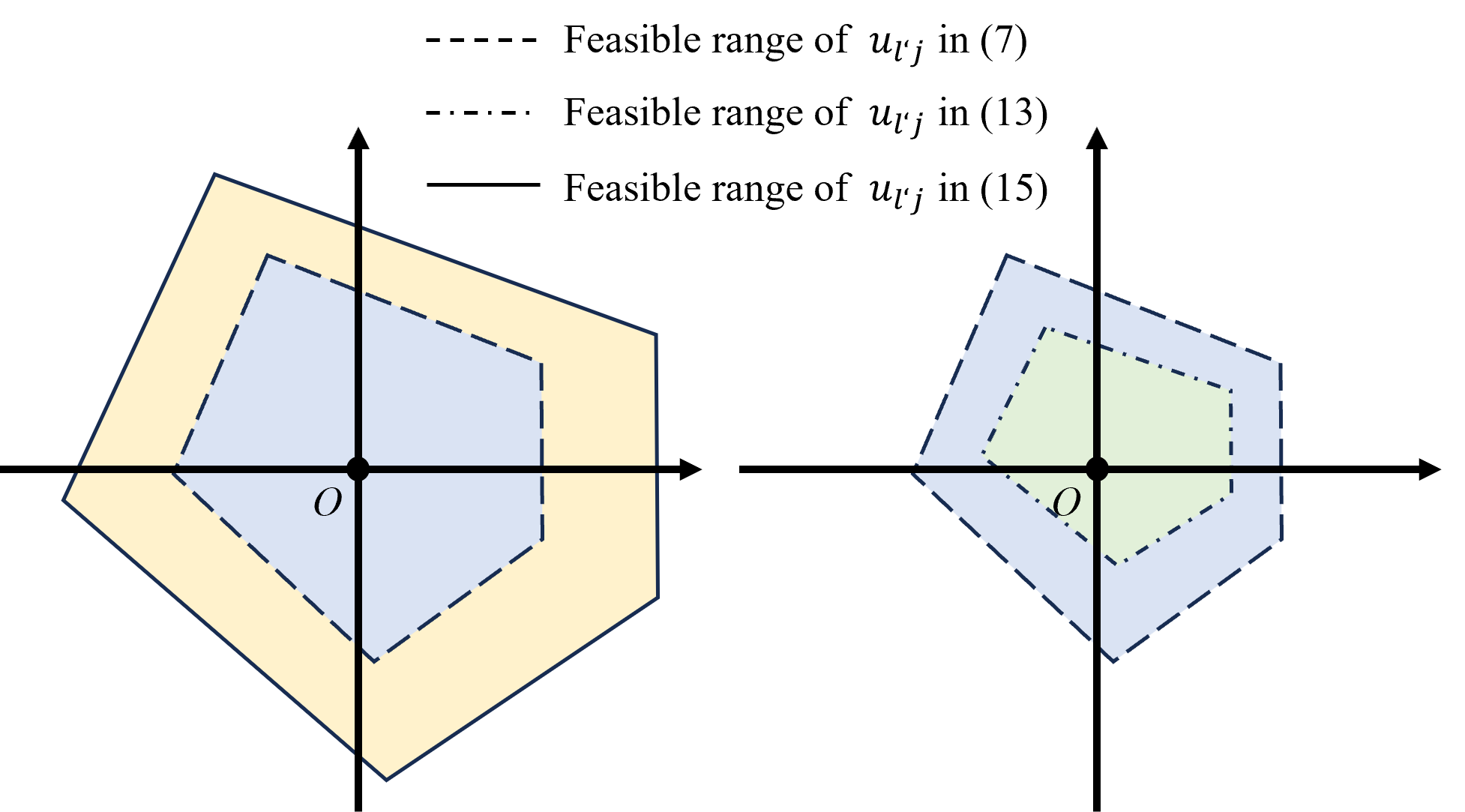} 
  \caption{Schematic illustration of the maximal and minimal control input range $u_{l'j}$.}
  \label{fig:predictive_reachability}
\end{figure}

The proof of Proposition~\ref{prop:2} follows the same reasoning as that of Proposition~\ref{prop:1} and is omitted for brevity. Fig.~\ref{fig:predictive_reachability} roughly illustrates the maximal expansion and contraction of the ranges of the control input $u_{l'i}$, determined by the uncertainty bound $\varepsilon_A, \varepsilon_B, \varepsilon_c$ under the unknown dynamics $\bar A_{l'}, \bar B_{l'}, \bar c_{l'}$, compared to the range generated by known $\bar A_l, \bar B_l, \bar c_l$.
\subsection{Adaptive Non-uniform State Space partitioning}\label{subsec:partitioning}
For a system with a state space of dimension $n$, a uniform partitioning scheme results in the number of discrete polytopes growing exponentially with $n$. Such discretization is computationally expensive when recalculating facet reachability and unreachability as the agent enters a new region. More importantly, this uniform treatment is unnecessary, as different areas of the state space typically require different levels of attention. For instance, regions around the agent’s current state or along its most probable future trajectories require more attention and thus higher discretization density, and otherwise requires less attention. In this subsection, we assume that the agent's state space $\mathcal{X}$ is a cube, $\mathcal{X}=\{x\in \mathbb{R}^n|x_i^- \leq x_i \leq x_i^+, i=1\ldots n\}$, and develop a non-uniform partitioning algorithm that adaptively refines the state space. The algorithm generates finer partitions in regions of high importance and coarser partitions elsewhere, thereby greatly reducing the total number of discrete polytopes while balancing between computational efficiency and local resolution.

Let $x_{c}$ and $x^*$ be the current state and the target state, respectively. At each iteration, a line segment $\overline{x_cx^*}$ is drawn between the two points, and all grid cells intersected by this line are identified. This segment represents the shortest Euclidean path between the current state and goal state and serves as a nominal direction of progress in the absence of detailed dynamic information. All the intersected cells are then subdivided uniformly into $2^n$ smaller cells. After the subdivision, the line segment is again checked for intersections with the newly generated cells, and the process is repeated. The refinement continues iteratively until all intersected cells reach the predefined minimum size, at which point the partitioning process terminates. We use cubes to implement the partition since cubes are simple to represent and partition equally~\cite{Sinclair2019Adaptive, ORNIK20179089}. However, we keep the option of partitioning the state space using any polytopes such as simplices, Voronoi partitions, etc~\cite{hoerger2024adaptive}.

The algorithm adaptively refines the state space partition along the segment connecting the current and goal states since feasible trajectories are likely to be close to this segment. Refining the partition along this segment concentrates computations on the most relevant region in the state space. Moreover, it ensures that the finest partitions are generated near both the current and goal states --- an intuitive outcome, as these regions naturally warrant greater attention during planning and control. 

\begin{figure}[htbp]
    \centering
    \begin{minipage}{0.234\textwidth}
        \centering
        \includegraphics[width=\textwidth]{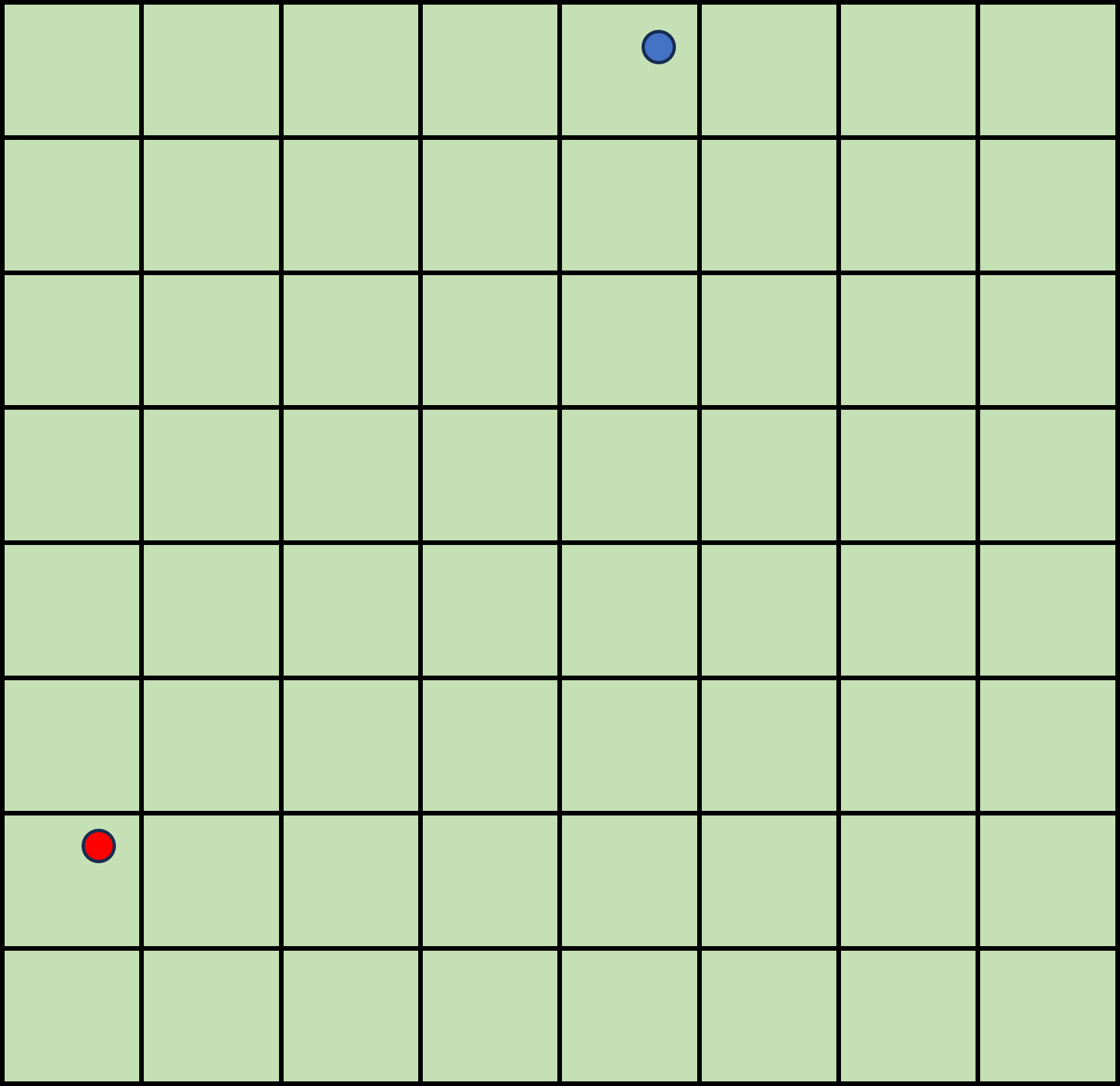}
    \end{minipage}
    \hfill
    \begin{minipage}{0.234\textwidth}
        \centering
        \includegraphics[width=\textwidth]{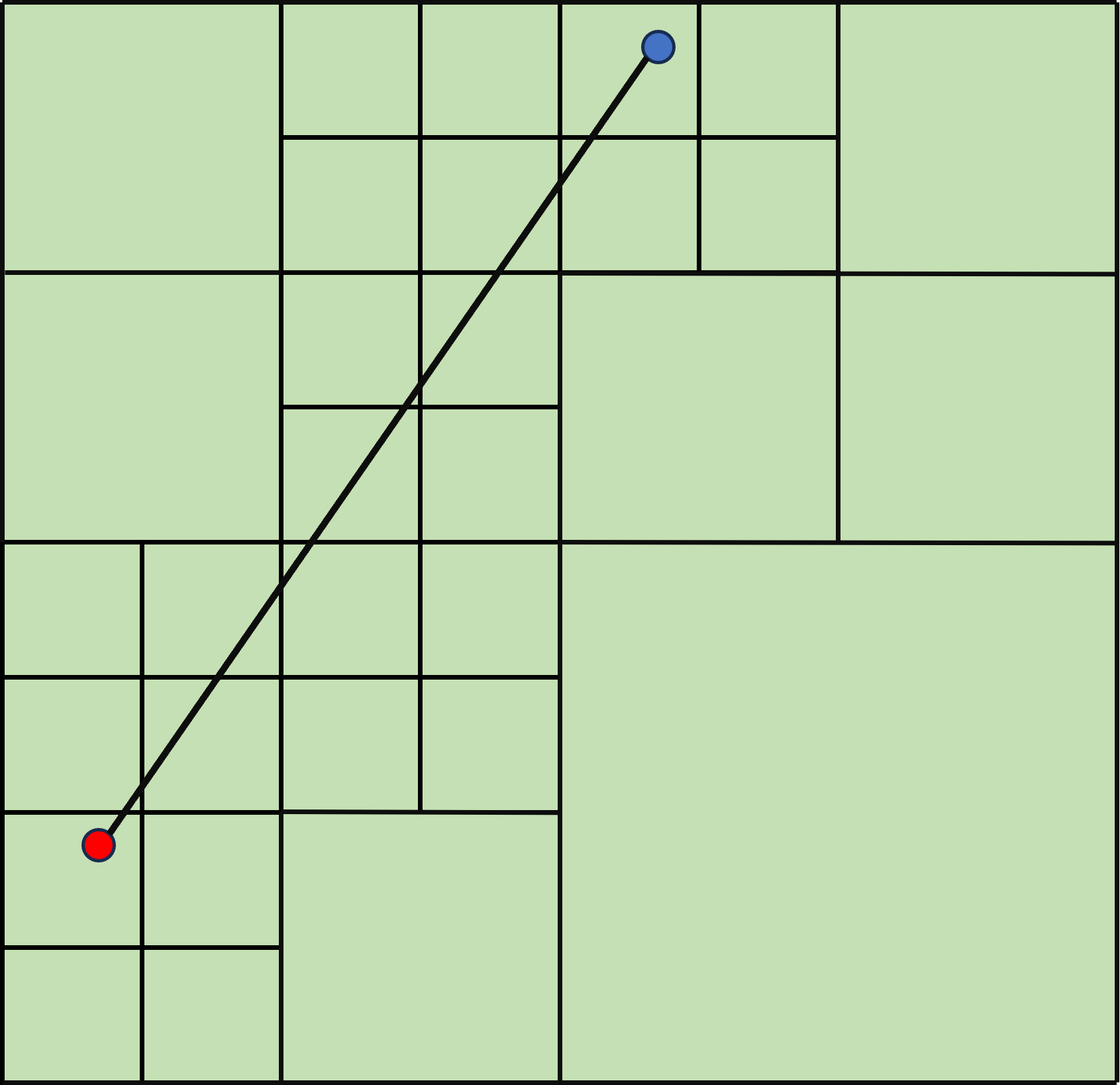}
    \end{minipage}
    \caption{A 2D example illustrates uniform and non-uniform state space partitioning. The blue dot represents the current state and the red dot represents the target state. (Left) Uniform partitioning. (Right) Non-uniform partitioning.}
    \label{fig:two_side_by_side}
\end{figure}

The detailed steps of the proposed procedure are presented in Algorithm~\ref{alg:adaptive-partition}, and an illustrative example is provided in Fig.~\ref{fig:two_side_by_side}.


\begin{algorithm}[h]
\caption{NonuniformPartitioning}
\begin{algorithmic}[1]\label{alg:adaptive-partition}
\REQUIRE Current state $x_c$, goal state $x^*$, root cube $C_{\mathrm{root}}\subset\mathbb{R}^n$, minimum cube size $h_{\min}$
\STATE Initialize the cube set $\mathcal{C} \gets \{C_{\mathrm{root}}\}$
\STATE Define the line segment $L = \overline{x_cx^*}$
\REPEAT
    \STATE $\mathcal{C}_{\mathrm{refine}} \gets \{\, C \in \mathcal{C} \mid L \cap C \neq \varnothing \text{ and } \mathrm{side}(C) > h_{\min} \,\}$
    \FOR{each cube $C \in \mathcal{C}_{\mathrm{refine}}$}
        \STATE Remove $C$ from $\mathcal{C}$
        \STATE Subdivide $C$ equally along each axis into $2^n$ sub-cubes
        \STATE Add the resulting $2^n$ sub-cubes to $\mathcal{C}$
    \ENDFOR
\UNTIL{$\mathcal{C}_{\mathrm{refine}} = \varnothing$}
\STATE \textbf{Output:} Final cube set $\mathcal{C}$
\end{algorithmic}
\end{algorithm}

\subsection{Predictive Reachability-guaranteed Graph}\label{subsec:graph}

We now describe how the high-level planning graph is constructed and weighted based on the predicted reachability analysis. The graph $G_s=(V_s, E_s, w_s)$ is built incrementally using the sufficient conditions established in Proposition~\ref{prop:1} and Proposition~\ref{prop:2}, which allow reachability to be certified or ruled out even when local dynamics are unknown. The construction also uses the adaptive partitioning in the previous, which continuously refines the graph.

Each vertex in $V_s$ corresponds to a polytope in the state space partition. A directed edge from $P_l$ to a neighboring polytope $P_{l'}$ is added only if the two polytopes share a common facet $F_{l,l'}$ and the robust feasibility condition \eqref{eq:narrow-ineq-new} holds when $F_{l,l'}$ is treated as the exit facet. On the other hand, if either the two polytopes are nonadjacent or the expanded inequality set \eqref{eq:expand-ineq-new} is infeasible, the transition from $P_l$ to $P_{l'}$ is excluded from the graph, as reachability can be conclusively ruled out. In particular, when the area of the outgoing facet of a polytope exceeds that of the incoming facets, the reachability of the corresponding transitions cannot be determined. In such cases, the associated edges in the reachability graph are classified as uncertain.

For transitions whose existence is guaranteed, edge weights are chosen to reflect the time required for the agent to exit the current polytope through the corresponding facet. 
\begin{pro}
    Consider a polytope $P_l$ with vertices $v_{l1}, \ldots, v_{lM}$ and associated control inputs $u_{l1}, \ldots u_{lM}$. Suppose that the agent enters $P_l$ at state $x_{l0}$ and exits through the facet with outward normal $n_{l1}$. An upper bound on the exit time $T_0$ for polytopes with certain dynamics can be obtained as 
    \begin{equation}
        T_0\leq \frac{\beta  - \alpha}{c_1},
    \end{equation}
    where $\alpha = \min\{n_1^\top v_j|j = 1,\ldots,M\}$, $\beta = \max\{n_1^\top v_j | 1, \ldots, M\}$, and $c_1 = \min\{n_1^\top (Av_j + Bu_j + c)|j=1,\ldots,M\}$.
\end{pro}

 See Appendix~C for a detailed derivation of this time bound and a robust time bound given that the affine dynamics are unknown but bounded. This bound follows directly from standard reach control arguments: a detailed derivation can be found in \cite{habets2004control}. For polytope $P_{l'}$ whose vertices are denoted by $v_{l'1},\ldots,v_{l'M}$ with unknown dynamics but certain facet reachability, we can also obtain a conservative upper bound on the exit time $T_0'$ via the bounded deviation of unknown dynamics and the identified dynamics. The resulting values are used as the weight for edges whose reachability is confirmed.

In contrast, when the conservative condition \eqref{eq:narrow-ineq-new} fails but the expanded condition \eqref{eq:expand-ineq-new} remains feasible, the reachability of the corresponding facet cannot be definitely verified. Rather than discarding such transitions, we retain them in the graph and assign them heuristic weights that encode their exploratory relevance. These weights are designed to reflect the expected reduction in uncertainty that would result if the agent traverses such an edge. Namely, we suppose that the prior information about the existence of an edge $e$ is available in the form of a probability $p_e$. 
\begin{rem}
If no prior information is available, we adopt the probability $p_e$ to be $p_e = \frac{1}{2}$.    
\end{rem}

The uncertainty associated with this edge is quantified using Shannon entropy~\cite{shannon1948mathematical}:
\begin{equation}
    H(e) = -[p_e\log(p_e)+(1-p_e)\log(1-p_e)].
\end{equation}
Accordingly, assuming that the existence of unknown edges is independent, the total uncertainty of the graph $G$ is defined as
\begin{equation}
H(G) = \sum_{e\in E}H(e).
\end{equation}
This provides only an approximation of the total uncertainty, since the existence of unknown edges is not independent, as they are coupled through the Lipschitz assumption. The \textit{information gain} associated with edge $e$ is defined as the reduction in graph uncertainty before and after traversing the edge, namely $IG_e = H(G)_{\rm before} - H(G)_{\rm after}$. If the edge exists, which is assumed to occur with probability $p_e$, the agent reaches the adjacent vertex and can subsequently determine at least the existence of the outgoing edges $e_i' \in  E_v$. 

Consequently, with probability $p_e$, the information gain also includes the sum of entropies associated with these outgoing edges. Taking expectation with respect to the existence of $e$, the expected information gain is given by
\begin{equation}
    E[IG_e] = p_e (\sum_{e_i' \in E_v}H(e_i')).
\end{equation}
We assign weights to uncertain edges by jointly accounting for their expected information gain and the size of the corresponding cell, which reflects the traversal distance required to cross the cell, according to
\begin{equation}
    w(e) = \frac{C_{u}\cdot l_{u}}{1+\beta_u E[IG_e]},
\end{equation}
where $C_{u}$ is a sufficiently large positive constant, $l_u$ denotes the length of the corresponding cell along the transition direction, and $\beta_u$ is a tunable positive parameter that tunes the degree of exploration.

We emphasize that this weighting strategy differs fundamentally from the scheme proposed in \cite{11245807}. While the latter only encourages exploration of regions that are distant from the explored regions geometrically, the novel proposed formulation explicitly models uncertainty using information-theoretic quantities and drives the planner toward transitions that are expected to yield greater uncertainty reduction. As a result, the exploration behavior is more interpretable and directly aligned with the objective of information acquisition. This strategy is particularly advantageous when uncertainty is not distributed according to geometric distance from explored and unexplored regions.

We now present the motion planning pipeline that unifies local dynamics identification, predictive reachability analysis, construction of graph $G_s$, motion planning on the graph, and corresponding controller synthesis.

\subsection{Motion Planning Framework}\label{subsec:framework}
Algorithm \ref{alg:motionplan} outlines the proposed motion planning framework. The procedure starts by applying the adaptive state space partitioning strategy described in Algorithm \ref{alg:adaptive-partition}, which induces a predictive reachability graph $G_s$ whose nodes correspond to polytopic regions, with initial and target nodes. Planning then proceeds through an iterative loop that couples high-level graph-based reasoning with low-level controller synthesis. At each iteration, the state space is further refined in a non-uniform manner, and the local system dynamics within the current polytope are identified using Algorithm~\ref{alg:sysid}, yielding an affine approximation of the underlying unknown dynamics. Using this local model together with prior bounds on model dynamics, facet reachability from the current polytope to its neighboring regions is evaluated, including transitions associated with polytopes whose dynamics have not yet been identified. 

The reachability graph $G_s$ is subsequently updated by pruning infeasible transitions, confirming feasible ones, and assigning weights to edges according to their status: verified edges are weighted using time-bound estimates, while uncertain edges are assigned heuristic weights reflecting their exploratory value. Given the updated graph, Dijkstra’s algorithm \cite{dijskstra, cormen2022introduction} is employed to compute a shortest path from the current node to the target node. The first node along this path, denoted by $Path(1)$, is selected as the next navigation objective. An affine feedback controller is then synthesized following the reach control procedure in \cite{habets2004control} to drive the agent toward the corresponding polytope. This combined exploration–navigation process repeats until the agent enters the target polytope.

It is important to note that the use of locally linearized affine models may occasionally steer the agent into regions that deviate from those predicted by the abstraction. Furthermore, since the true system dynamics are unknown and global reachability cannot be guaranteed even under known dynamics, situations may arise in which the agent becomes confined within a polytope or fails to progress toward the target. These scenarios motivate the need for continual model updating and adaptive planning within the proposed framework.

\begin{algorithm}
\caption{Hierarchical Motion Planning and Control Framework}
\begin{algorithmic}[1]\label{alg:motionplan}
\REQUIRE Lipschitz Constants $L_{df},L_g$, minimum tile size $h_{\min}$, exploration parameters $C_u, p_e, \beta_u$, initial state $x_{\rm initial}$, target state $x_{\rm target}$

\STATE \textbf{Initialize:}
\[
\begin{array}{l}
 Node_{\text{initial}}, Node_{\text{target}} \gets \text{NonuniformPartitioning}\\
 \ \ \ \ \text{($x_{\rm initial}$, $x_{\rm target}$)}\\
 Node_{\text{current}} \gets Node_{\text{initial}}\\
 Node_{\text{explored}} \gets \text{empty list}\\
 x_{\rm current} \gets x_{\rm initial}\\
\end{array}
\]
\IF{$Node_{\text{current}} \neq Node_{\text{target}}$}
\STATE \parbox{1\linewidth}{
$Node_{\rm current}, Node_{\rm target} \gets \text{NonuniformPartitioning}$\\
$\text{\ \ \ \ }(x_{\rm current}, x_{\rm target})$
}

    \STATE $[A,B,c] \gets \text{AffineSystemIdentification}\ (Node_{\text{current}})$
    \STATE Append $Node_{\text{current}}$ to $Node_{\text{explored}}$
    \STATE $\text{Construct $G_s$}\ (A, B, c, L_{df}, L_g)$
    \STATE AssignTimeBoundWeights $(G_s,A,B,c,L_{df}, L_g)$ \COMMENT{for certain edges}
    \STATE AssignInfoGainWeights $(G_s, C_u, p_e, \beta)$ \COMMENT{for uncertain edges}
    \STATE $\text{Path} \gets \text{Graph Search ($G_s$, $Node_{\text{current}}$, $Node_{\text{target}})$}$
    \STATE $Node_{\text{current}} \gets \text{Path(1)}$
    \STATE Synthesize Controller to Reach $Node_{\text{current}}$
    \STATE Update $x_{\rm current}$
\ELSE
    \STATE apply CLF-CBF controller to drive the robot to $x_{\rm target}$ inside $Node_{\rm target}$
\ENDIF
\end{algorithmic}
\end{algorithm}

\section{Case studies}\label{sec:case}
In this section, we present two case studies involving a fully actuated mobile robot and an underactuated mobile robot. For the underactuated case, we introduce the necessary relaxations and modifications to the basic theoretical framework to appropriately accommodate the lack of full actuation.
\subsection{Fully-actuated Mobile Robot}
\begin{figure*}[t]
  \centering
  \includegraphics[width=\textwidth]{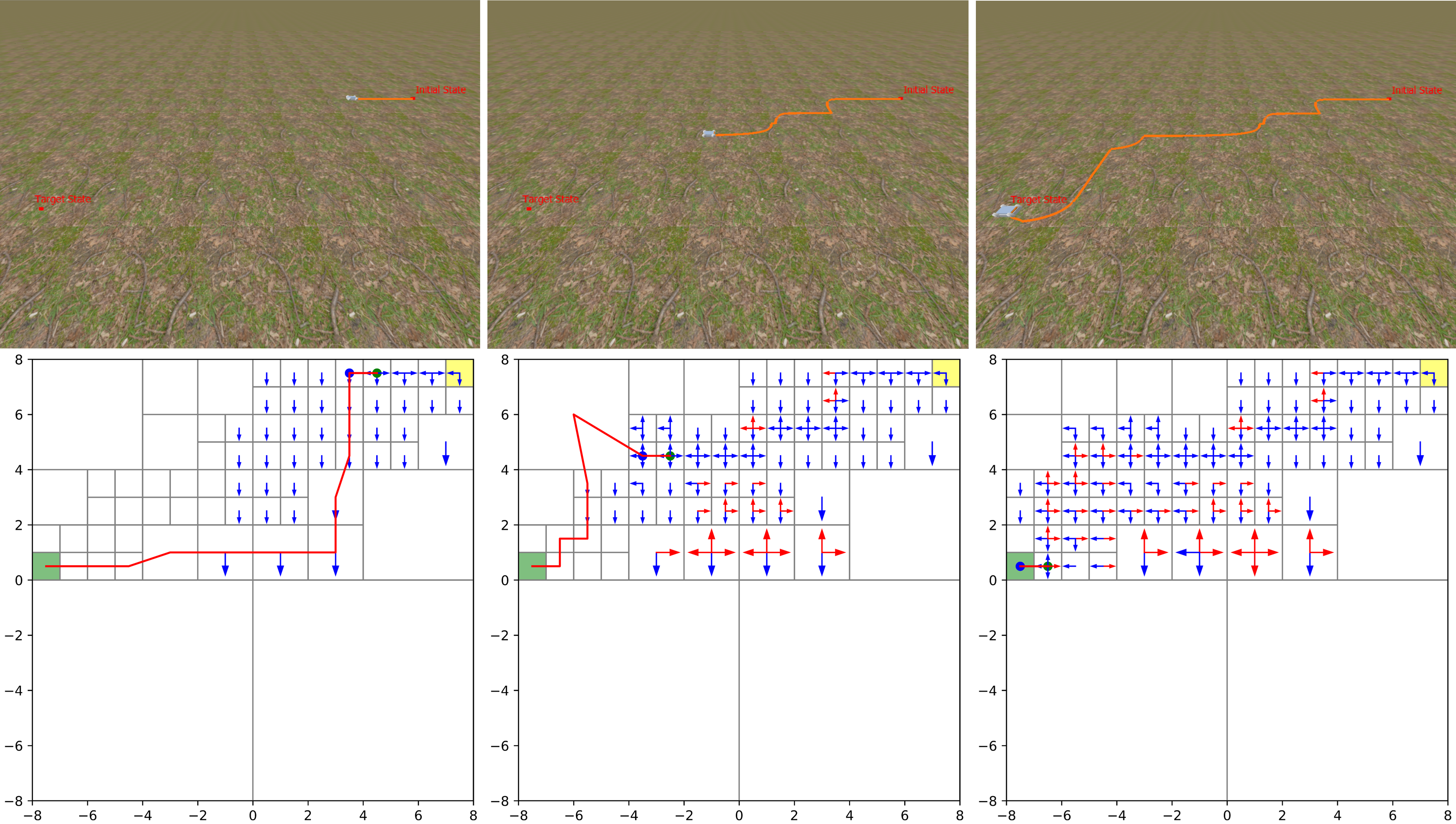} 
  \caption{Simulation results for a fully actuated mobile robot operating under unknown dynamics. Each column corresponds to a distinct time instant during the simulation. The top row illustrates the robot’s trajectory within the simulated environment, while the bottom row depicts the associated reachability analysis and prediction, nonuniform adaptive state space partitioning, and high-level planning performed on the reachability-guaranteed graph. Blue arrows indicate facets that are provably reachable, red arrows denote the absence of feasible transitions, and the absence of an arrow represents uncertainty in edge existence. The initial tile is highlighted in yellow and the target tile is highlighted in green. The green dot and blue dot denote the current tile and the next tile on the predicted graph respectively. The red line shows the optimal path computed at each discrete planning step.}
  \label{fig:1}
\end{figure*}

The true dynamics of a four-wheeled mobile robot equipped with Mecanum wheels~\cite{macanum} and operating on unknown terrains are modeled in the planar workspace. Since Mecanum wheels enable holonomic motion in the plane, we treat the robot as fully actuated in translational directions. In this case study, the robot’s yaw angle is kept constant, and we focus on the translational state $x = [x_1, x_2]^\top$, which corresponds to the robot $x-$ and $y-$position on the ground plane. The resulting closed-loop translational behavior—affected by wheel–ground interaction, local slip, and terrain-induced bias—is represented by the following control-affine model~\cite{11245807}:
\begin{equation}
    \begin{aligned}
        \dot x &= f(x) + g(x)u,\\
        f(x) &= \begin{bmatrix}
            -0.5 \sin(0.1x_1 - 0.2x_2)-4.5\\
            -0.2\sin(0.3x_1-0.1x_2)-4.5
        \end{bmatrix},\\
        g(x) &= \begin{bmatrix}
            1+0.02x_1 & 0.02x_2\\
            -0.02x_1 & 1-0.02x_2
        \end{bmatrix}.
    \end{aligned}
\end{equation}
Here, $f(x)$ captures terrain-dependent drift effects (e.g., slope-induced bias and unmodeled wheel slip), while $g(x)$ describes a state-dependent input mapping that accounts for variations in effective traction and coupling between the commanded planar velocities and the realized motion. The functions $f(x)$ and $g(x)$ are treated as unknown to the agent, except that the Lipschitz constants are assumed to be known as $L_{df} = 0.03$ and $L_g = 0.03$. These constants are more conservative than those of the real dynamics. 

The planar state space is constrained by $-8\ {\rm m}\leq x_1 \leq 8\ {\rm m}$ and $-8\ {\rm m}\leq x_2 \leq 8\ {\rm m}$, which defines the workspace used for evaluating the proposed method. The two-dimensional control input $u \in \mathbb{R}^2$ is bounded by $-5\ {\rm m/s} \leq u_i \leq 5\ {\rm m/s},\ i=1,2$, representing bounded planar velocity commands applied to the robot. These bounds are chosen to provide a reasonable operating range for the simulation study. We employ the proposed adaptive nonuniform state space partitioning: starting from a root square covering the entire domain, we iteratively refine the cells intersected by the line segment connecting the current state to the target state. Each refinement splits a selected cell into 4 subcells, and the refinement terminates once the minimum cell side length reaches 1. The parameters of unknown edges are set to $C_u = 100$ and $\beta = 0.8$, making these parameters sufficiently large (larger than those of certain edges). The simulation shows that the computed facet reachability involves not only grid cells with identified dynamics, but also predicts facet reachability in regions with unknown dynamics around the explored region. At the end of the simulation scenario, our non-uniform adaptive state space partition algorithm resulted in a $68\%$ reduction in the number of split tiles compared to a uniform partition, which significantly decreases the computational cost of constructing the reachability-guaranteed graph.
The simulation result is shown in Fig.~\ref{fig:1}. It demonstrates that our algorithm identifies a feasible path from the initial to the target and synthesized corresponding controllers to realize the planning.

\begin{figure*}[t]
  \centering
  \includegraphics[width=\textwidth]{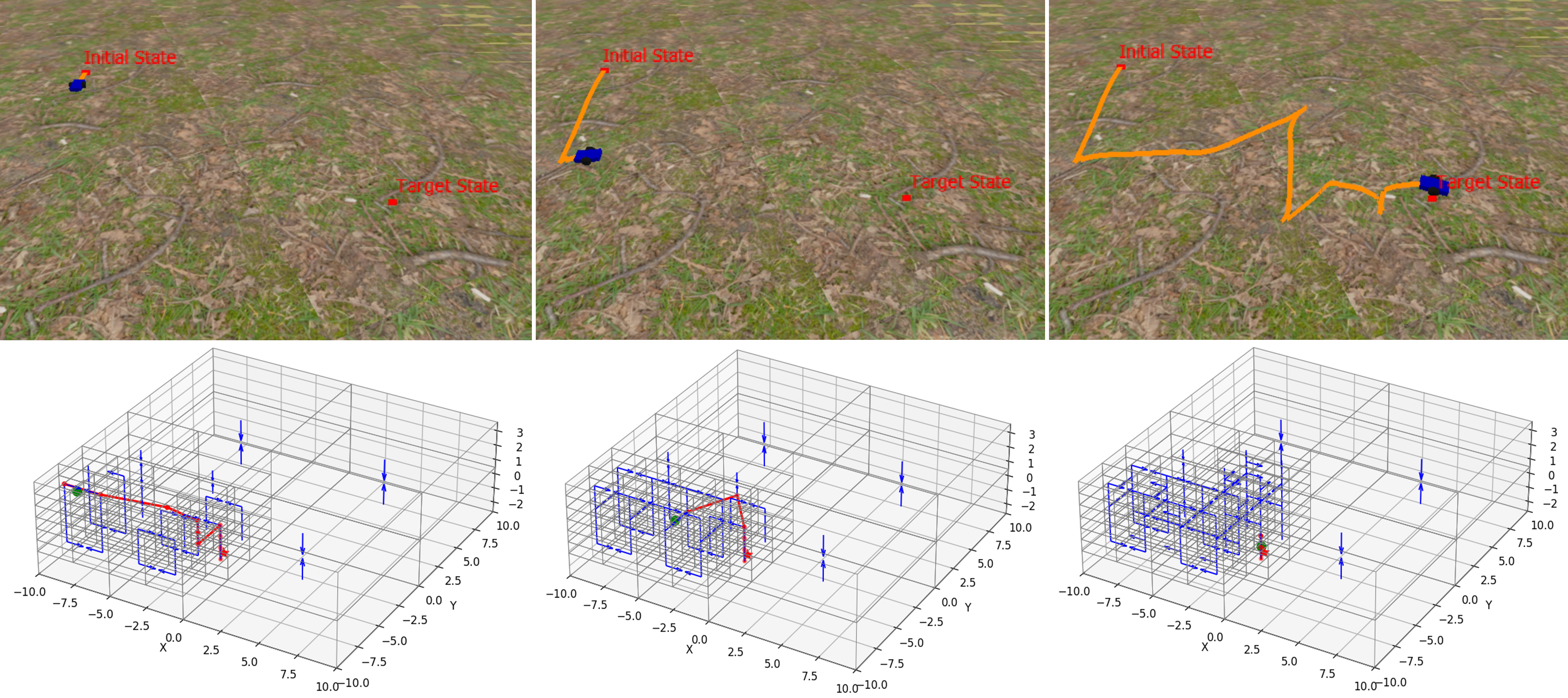} 
  \caption{Simulation results for an under-actuated mobile robot operating under unknown dynamics. Each column corresponds to a distinct time during the simulation. The top row illustrates the robot's trajectory (marked as orange line) in the simulated environment. The bottom row depicts the associated reachability analysis and prediction, nonuniform adaptive state space partitioning, and high-level planning performed on the reachability-guaranteed graph. Blue arrows indicate facets that are reachable. The target tile is marked by a red star. The red line represents the sequence of polytopes selected by the high-level planner, and the green dot denotes the robot's current state.}
  \label{fig:222}
\end{figure*}
\subsection{Under-actuated Mobile Robot}
For underactuated systems, the control constraints imposed by classical reach control conditions can be overly restrictive, which often lead to infeasibility in the control synthesis. In this section, we introduce modifications to the reach control framework that enable its application to underactuated systems, followed by a case study of an underactuated mobile robot operating under unknown disturbances.

We consider the kinematics model of a unicycle-type mobile robot with disturbances \cite{sun2018robust}:

\begin{equation}
\begin{aligned}
    \begin{bmatrix}
        \dot x\\
        \dot y\\
        \dot \theta
    \end{bmatrix} &= \begin{bmatrix}
        \cos \theta & 0\\
        \sin \theta & 0\\
        0 & 1
    \end{bmatrix}\begin{bmatrix}
        v + d_v\\
        \omega + d_\omega
    \end{bmatrix} \\&= \underbrace{\begin{bmatrix}
        \cos \theta & 0\\
        \sin \theta & 0\\
        0 & 1
    \end{bmatrix}}_{g} \underbrace{\begin{bmatrix}
        v\\
        \omega
    \end{bmatrix}}_{u} + \underbrace{\begin{bmatrix}
        \cos\theta \cdot d_v\\
        \sin \theta \cdot d_v\\
        d_\omega
    \end{bmatrix}}_{f},
    \end{aligned}\label{eq:car_dynamics}
\end{equation}
where $x,y\in \mathbb{R}$ denote the position coordinates, $\theta \in [-\pi, \pi]$ denotes the orientation, $v,\omega \in[-10, 10]$ denote the linear and angular velocities respectively.  $d_v$ and $d_\omega$ denote state-variant unknown disturbances on linear and angular velocities. We firstly introduce relaxed facet reachability conditions so that reach control theory can be more applicable to underactuated systems.
\begin{figure}[htbp]   
    \centering
        \centering
        \includegraphics[width=0.4\textwidth]{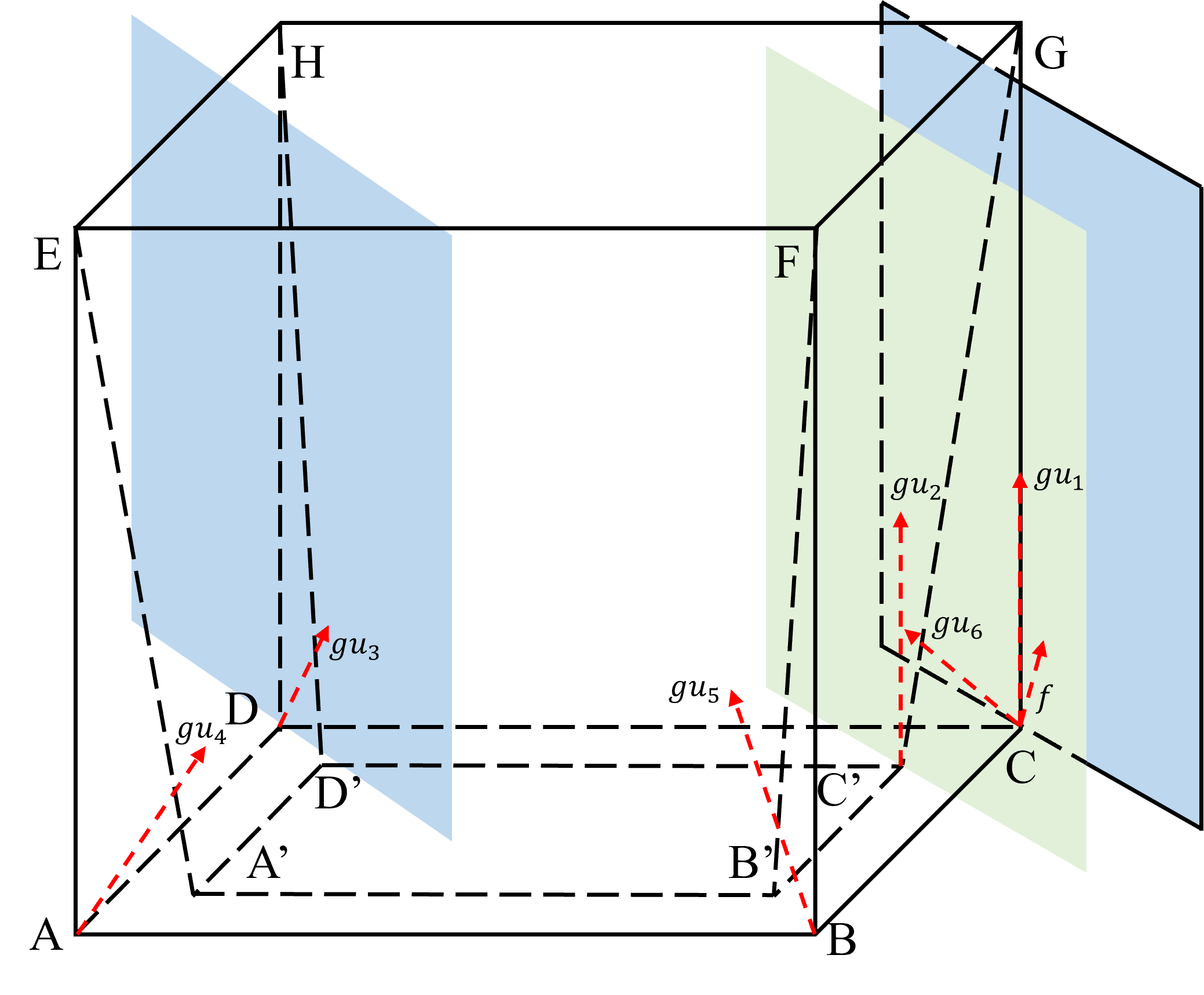}
    \caption{An example of a three-dimensional cube $ABCDEFGH$ and a subpolytope (truncated pyramid) $A'B'C'D'EFGH$. The blue planes represent the set of velocities generated by $gu$ of vertices $C$ and $D$, and the green plane represents the set of velocities generated by $gu$ of vertex $C'$.}
     \label{fig:relaxation}
\end{figure}
\begin{figure}[htbp]   
    \centering
        \centering
        \includegraphics[width=0.4\textwidth]{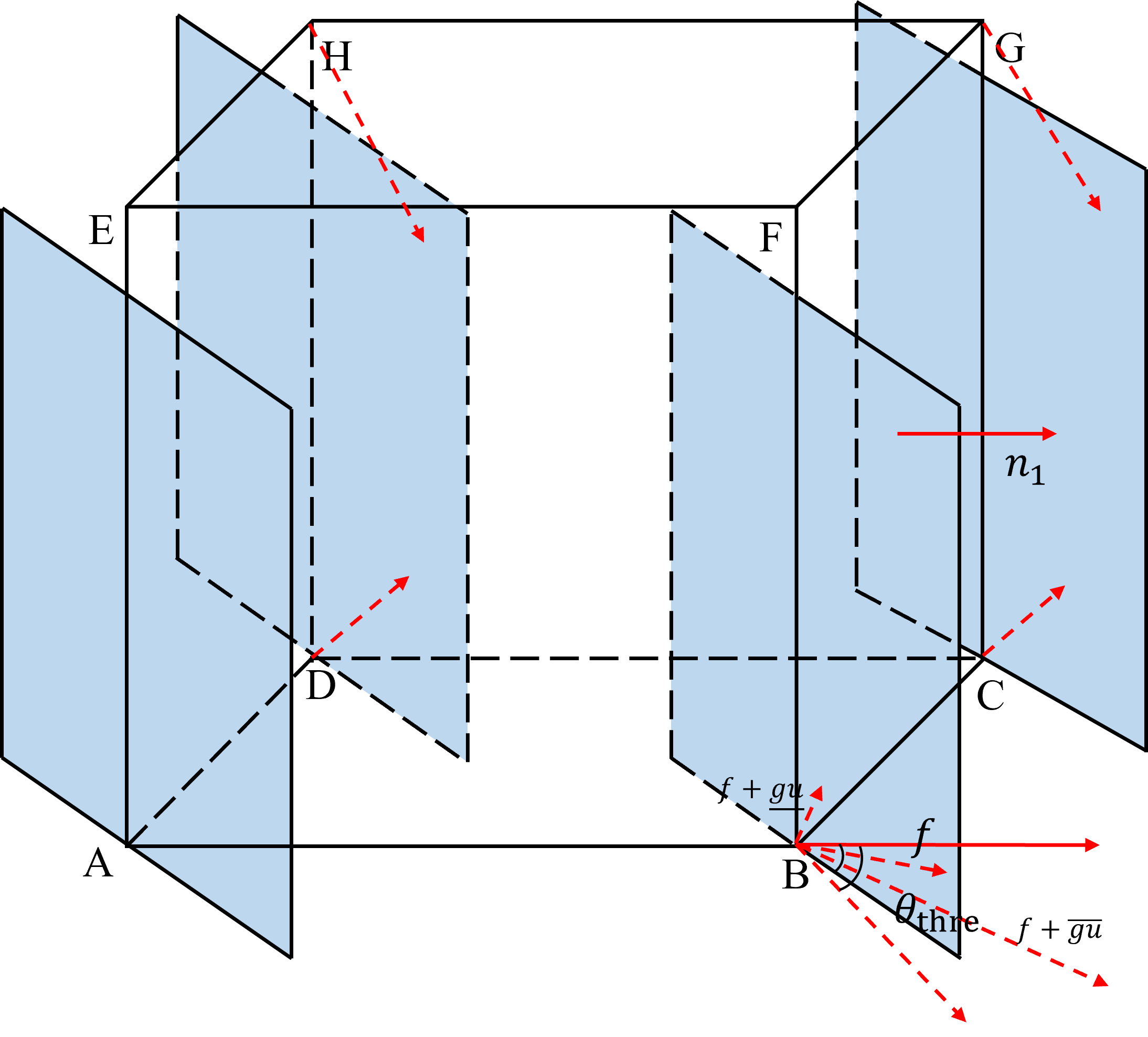}
    \caption{An illustration of relaxed facet reachability conditions on side facets.}
     \label{fig:relaxation1}
\end{figure}
\subsubsection{Relaxed Facet Reachability}
 In reach control theory, both facet reachability analysis and controller synthesis are based on identifying feasible control inputs at the vertices of polytopes in the state space. We provide an example to illustrate how classical reach control conditions can be restrictive. Consider a three-dimensional state space as shown in Fig.~\ref{fig:relaxation}. Suppose the state space is the cube ABCDEFGH with exit face EFGH and temporarily neglect the drift term $f$. The set of velocities that can be generated by the control term $gu$ is confined to a lower-dimensional affine subspace, depicted as the blue plane in Fig.~\ref{fig:relaxation}. In this case, the admissible velocity at each vertex must point inward to remain within the polytope, such as $gu_3$, $gu_4$, $gu_5$ and $gu_6$. For vertex $C$, for instance, the only admissible velocity is $gu_1$. As a result, when taking the drift term $f$ into consideration, it may be impossible to find control inputs that produce admissible velocities at all vertices $A$, $B$, $C$, and $D$. 

This limitation arises from the fact that underactuation restricts the admissible control directions at each vertex, which may prevent the reachability conditions from being satisfied. This motivates the need for relaxing the original polytope in order to recover feasibility conditions. We propose a relaxation strategy for a three-dimensional cube with control term $gu$ spans a plane perpendicular to the top and bottom facets. This strategy is categorized into two cases: facet reachability toward the top and bottom facets, and facet reachability toward the side facets.

\underline{\textbf{Top and bottom facets:}} To assess the reachability of the top facet $EFGH$, we introduce a subpolytope in the form of a truncated pyramid, denoted by $A'B'C'D'EFGH$, where the facet $A'B'C'D'$ is a scaled-down version of the original facet $ABCD$. Consider, for example, the vertex pair $C$ and its corresponding point $C'$. In this construction, the set of admissible control-induced velocities is no longer restricted to the blue plane ($gu_1$ for instance), but instead lies within the green plane ($gu_2$ for instance). This strategy significantly enlarges the feasible set of vertex-wise control inputs by allowing a broader range of directions pointing inward the truncated pyramid $A'B'C'D'EFGH$ within the green plane. An analogous construction is applied when evaluating the reachability of the bottom facet $ABCD$, where a pyramid-shaped subpolytope $E'F'G'H'ABCD$ is considered, with $E'F'G'H'$ being a reduced facet of $EFGH$. If the system state lies within the corresponding subpolytope and the relaxed reachability conditions are satisfied, facet reachability can be guaranteed. However, one should notice that when the state lies outside the subpolytope, no such guarantees can be established for the original cube. A similar approach is reported in~\cite{ORNIK20179089} where the exit facet is widened instead of narrowing the facet opposite to the exit facet.

\underline{\textbf{Side facets: }}We next introduce the relaxation strategy for evaluating reachability of the side facets. As in the previous case, we temporarily neglect the velocity contribution from the drift term and focus exclusively on the control-induced velocity $gu$. As illustrated in Fig.~\ref{fig:relaxation1}, the orientation of the admissible velocity plane depends on the system state $\theta$ in~\eqref{eq:car_dynamics}. Consider, for instance, the reachability of the side facet $BCFG$, which has the normal vector $n_1$. Under this configuration, feasible control inputs can be identified at vertices $C$, $G$, $D$, and $H$, whereas no admissible control signals exist at the remaining vertices. To address this issue, we introduce a threshold angle $\theta_{\rm thre}$. If for $\theta$ which corresponds to infeasible control inputs, and $|\theta|\leq \theta_{\rm thre}$, the control input associated with the corresponding vertex is still considered feasible under a relaxed criterion, and the control inputs at vertices where feasibility conditions cannot be achieved are set to zero when synthesizing controllers. We now take into account the drift term in~\eqref{eq:car_dynamics} and compute the \textit{limiting angles} formed between the vector field $f+gu$ and the normal vector of the target facet. Take vertex $B$ in Fig.~\ref{fig:relaxation1} for instance, the limiting angles are represented by the angle between $f+\underline{gu}$ and $\theta$, and the angle between $f+\overline{gu}$ and $\theta$. If the classical facet reachability conditions are not feasible with a vertex, but the absolute values of these corresponding angles are less than $\theta_{\rm thre}$, the corresponding vertex is considered feasible, and the control input is set to be zero when synthesizing controller. The controller synthesized by this strategy might drive the agent exit from the wrong facet for some initial points. However, by choosing $\theta_{\rm thre}$ sufficiently small, the set of such initial points can be small. This comes at the cost of making the reachability condition more restrictive.

\subsubsection{Simulation Results}
A simulation result for the underactuated robot is shown in Fig.~\ref{fig:222}. The unknown disturbances on the velocity and angular velocity are set to be $d_v = 0.03\cos(0.01x + 0.02y)\ {\rm m/s}$, $d_\omega = 0.03\sin(-0.02x + 0.01y)\ {\rm rad/s}$, which simulate position-dependent effects such as terrain irregularities and unmodeled wheel-ground interactions. The disturbance magnitudes are chosen to remain small relative to the nominal velocity commands while still introducing nontrivial model uncertainty. The state space is constrained to $-10\ {\rm m}\leq x\leq 10\ {\rm m}$, $-10\ {\rm m} \leq y \leq 10\ {\rm m}$, $-\pi\ {\rm rad} \leq \theta \leq \pi\ {\rm rad}$. The upper bound for each control signal component is 10° and the lower bound is -10°. The minimum grid size is $h_{\min}^x = h_{\min}^y = 1.25\ {\rm m}$, $h_{\min}^\theta = \frac{\pi}{4}\ {\rm rad}$. The parameters for the weights assigned to unknown edges are set to be $C_u = 10$ and $\beta = 1$. The margin for relaxation mentioned above is $\theta_{\rm thre} = 10\degree$. Our simulation shows that the under-actuated mobile robot is able to predict facet reachability of the tiles and synthesis corresponding controllers to execute the planning trajectory under our relaxation strategy of the reach control theory. Similar to the fully-actuated scenario, the non-uniform partition strategy reduces computational cost and the proposed algorithm is able to find a feasible path under unknown dynamics. The choice of $\theta_{\rm thre}$ influences both feasibility determination and the risk of exiting through an unintended facet. A larger value of $\theta_{\rm thre}$ relaxes the feasibility condition but increases the likelihood of transitioning through an unintended facet.

\section{Conclusion and Discussion}\label{sec:conclusion}
This paper proposed a hierarchical motion planning and control framework for robotic systems operating under unknown nonlinear dynamics. By integrating affine system identification, predictive reachability analysis, non-uniform state space partitioning and graph-based planning, the framework enables online navigation while balancing exploration and exploitation. The approach accommodates both fully-actuated and under-actuated systems through appropriate reachability relaxations. Simulation results demonstrate the effectiveness of our approach, verifying its capability for scalable and adaptive robotic motion planning.

The proposed framework relies on the accuracy of local affine system identification to ensure that the planning and control are consistent. In practice, modeling errors may lead to small discrepancies, such that the synthesized controller does not always steer the system to exit a polytope through the intended facet. This effect is more obvious for under-actuated system since relaxed reachability conditions are introduced. However, as the algorithm is planned online adaptively, the new feasible path will be found once the execution deviates from the planned transition. Furthermore, while a CLF-CBF-based controller is employed within the terminal polytope, no theoretical guarantee exists for the existence of a feasible feedback control that drives the system exactly to the target state. Nevertheless, entering the terminal polytope implies that the system is already close to the target.

\section*{Appendix A \\ Difference bounds between local affine models}
Consider two affine systems derived as in~\eqref{eq:PWA} by linearizing the nonlinear dynamics in \eqref{eq:ds} at two distinct operating points, $(x_1, u_e=0)$ and $(x_2, u_e=0)$. The resulting affine representations are denoted by $(A_1, B_1, c_1)$ and $(A_2, B_2, c_2)$ respectively. Our goal is to establish rigorous bounds on the differences between these local affine approximations. Specifically, we seek constants $\varepsilon_A$, $\varepsilon_B$, and $\varepsilon_c$ such that $\|A_2 - A_1\|\leq \varepsilon_A$, $\|B_2-B_1\|\leq \varepsilon_B$, $\|c_2-c_1\|\leq \varepsilon_c$.

From Assumption \ref{ass:Lip}, it follows directly that 
\begin{equation}
    \begin{aligned}
    \|A_2-A_1\| &=\|\nabla_x f(x_2) - \nabla_x f(x_1)\| \leq L_{df}\|x_2-x_1\|,\\
    \|B_2-B_1\| &= \|g(x_2)-g(x_1)\| \leq L_g \|x_2-x_1\|.
\end{aligned}
\end{equation}

Hence, the corresponding bounds can be expressed as $\varepsilon_A = L_{df}\|x_2-x_1\|$, $\varepsilon_B = L_g\|x_2-x_1\|$.

Next, we derive an upper bound on $\|c_2-c_1\|$, which captures the deviation in the affine offset term. The value of $\|c_2 - c_1\|$ can be bounded as
\begin{equation}
    \begin{aligned}
        \|c_2 - c_1\|=& \|(f(x_2)-f(x_1))-(A_2x_2-A_1x_1)\|\\
        =& \|(f(x_2)-f(x_1)-A_1(x_2-x_1))\\
        &-(A_2 - A_1)x_2\|\\
        \leq& \|f(x_2)-f(x_1)-A_1(x_2-x_1)\|\\
        &+ \|(A_2-A_1)x_2\|.
    \end{aligned}
\end{equation}
For term $\|(A_2-A_1)x_2\|$,
\begin{equation}
    \|(A_2-A_1)x_2\|\leq \|A_2-A_1\|\|x_2\|\leq (L_{df}\|x_2-x_1\|)\|x_2\|.
\end{equation}
For term $\|f(x_2) - f(x_1)-A_1(x_2-x_1)\|$, we firstly derive $f(x_2) - f(x_1)$. We define an auxiliary function $\phi(t) = f(x_1 + t(x_2 - x_1))$, thus $f(x_2) - f(x_1) = \phi(1) - \phi(0) = \int_0^1 \frac{d}{dt}\phi'(t)dt$, and $\frac{d}{dt}\phi(t) = \nabla_x f(x_1 + t(x_2 - x_1))(x_2 - x_1)$.
Consequently,
\begin{equation}
\scalebox{0.96}{$\begin{aligned}
        f(x_2)-&f(x_1)-A_1(x_2-x_1) \\
        =& \int_0^1 \nabla_xf(x_1+t(x_2-x_1))(x_2-x_1)dt\\& - \nabla_x f(x_1)(x_2-x_1)\\
    =& (\int_0^1 [\nabla_x f(x_1 + t(x_2-x_1))-\nabla_x f(x_1)]dt) (x_2-x_1).
\end{aligned}$}
\end{equation}
Taking its norm, we can obtain
\begin{equation}
    \begin{aligned}
        \|f(x_2)-&f(x_1)-A_1(x_2-x_1)\| \\\leq& \|\int_0^1 [\nabla_x f(x_1 + t(x_2-x_1))-\nabla_x f(x_1)]dt\|\\ &\|x_2-x_1\|\\
        \leq & (\int_0^1 \|\nabla_x f(x_1 + t(x_2-x_1))-\nabla_x f(x_1)\|dt) \\&\|x_2-x_1\|.
    \end{aligned}
\end{equation}
From the Lipschitz condition, we have
\begin{equation}
\begin{aligned}
    \|\nabla_x f(x_1+t(x_2-x_1))&-\nabla_x f(x_1)\| \\\leq& L_{df}\|(x_1+t(x_2-x_1))-x_1\|\\
    =&L_{df}t\|x_2-x_1\|.
\end{aligned}
\end{equation}
Then, 
\begin{equation}
    \begin{aligned}
         \|f(x_2) - f(x_1) &- A_1(x_2 - x_1)\| \\&\leq (\int_0^1 t L_{df}\|x_2 - x_1\|  dt)\|x_2 - x_1\|\\
        & = \frac{1}{2} L_{df}\|x_2 - x_1\|^2.
    \end{aligned}
\end{equation}
Thus, $\|c_2-c_1\|$ is bounded by
\begin{equation}
    \|c_2-c_1\| \leq \frac{1}{2}L_{df}\|x_2-x_1\|^2 + L_{df}\|x_2-x_1\|\|x_2\|=\varepsilon_c.
\end{equation}

\section*{Appendix B \\ Proof of Proposition 1}

For $i \in I_+$, the constraints about feasible control signals on vertices $v_{l'j}$ are represented by $n_{l' i}^\top (\bar A_{l'}v_{l' j} + \bar B_{l'}u_{l'j} + \bar c_{l'})>0$. We seek to provide a lower bound on $n_{l' i}^\top (\bar A_{l'}v_{l' j} + \bar B_{l'}u_{l'j} + \bar c_{l'})$ with known $(\bar A_l, \bar B_l, \bar c_l)$ and dynamics difference bounds expressed by $\|\bar A_l - \bar A_{l'}\|\leq \varepsilon_A$, $\|\bar B_l - \bar B_{l'}\|\leq \varepsilon_B$, $\|\bar c_l - \bar c_{l'}\|\leq \varepsilon_c$. For term $n_{l'i}^\top (\bar A_{l'}v_{l'j} + \bar c_{l'})$,
\begin{equation}
    \scalebox{0.93}{$\begin{aligned} 
           n_{l'i}^\top& (\bar A_{l'}v_{l'j} + \bar c_{l'}) = n_{l'i}^\top [(\bar A_l+\bar A_{l'} - \bar A_l)v_{l'j} + (\bar c_{l}+\bar c_{l'}-\bar c_l)]\\
        &\leq n_{l'i}^\top (\bar A_l v_{l' j}+\bar c_l) + \|n_{l'i}\|(\|\bar A_{l'}-\bar A_l\|\|v_{l'j}\|+\|\bar c_{l'}-\bar c_l\|)\\
        &\leq n_{l'i}^\top (\bar A_l v_{l' j}+\bar c_l) + \|n_{l'i}\|(\varepsilon_A \|v_{l'j}\|+\varepsilon_c)
    \end{aligned}$}
\end{equation}
For term $n_{l'i}^\top \bar B_{l'} u_{l'j}$,
\begin{equation}
    n_{l'i}^\top \bar B_{l'} u_{l'j} = \begin{bmatrix}
        n_{l'i}^\top\bar B_{l'}^1\ \cdots\ n_{l'i}^\top\bar B_{l'}^m
    \end{bmatrix}\begin{bmatrix}
        u_{l'j}^1\\
        \vdots\\
        u_{l'j}^m
    \end{bmatrix},
\end{equation}
where $\bar B_{l'}^i,i=1\ldots m$, represents the $i$-th column of $\bar B_{l'}$. If $u_{l'j}^i>0$,
\begin{equation}
    \begin{aligned}
        n_{l'i}^\top\bar B_{l'}^iu_{l'j}^i& = n_{l'i}^\top(\bar B_{l}^i + \bar B_{l'}^i-\bar B_{l}^i) u_{l'j}^i\\
        &\ge (n_{l'i}^\top\bar B_{l}^i - \|n_{l'i}^\top\| \|\bar B_{l}^i-\bar B_{l'}^i\| )u_{l'j}^i\\
        &\ge (n_{l'i}^\top\bar B_{l}^i - \|n_{l'i}^\top\| \varepsilon_B )u_{l'j}^i.
    \end{aligned}
\end{equation}
Conversely, if $u_{l'j}^i\leq0$, $n_{l'i}^\top\bar B_{l'}^iu_{l'j}^i \ge (n_{l'i}^\top\bar B_{l}^i + \|n_{l'i}^\top\| \varepsilon_B )u_{l'j}^i$. Combining the above, we obtain the following expression:
\begin{equation}
        \scalebox{0.87}{$\begin{aligned} 
 \bar B_{l'}u_{l'j} \ge  \left\{n^\top_{l'i}\bar B_{l}-\left[\textbf{sgn}(u_{l'j}^1)\varepsilon_B\|n_{l'i}\|\ \ldots\ \textbf{sgn}(u_{l'j}^1)\varepsilon_B\|n_{l'i}\|\right ] \right\}u_{l'j},
    \end{aligned}$}
\end{equation}
where we use $\textbf{sgn}(u_{l'j}^i)$ to denote the sign of $u_{l'j}^i$ and ``$>$'' means element-wise comparison. Hence, if 
\begin{equation}
    \bar B_{l'}^-u_{l'j}> -n_{l'i}^\top \bar A_{l} v_{l'j} - n_{l'i}^\top \bar c_{l} + \|n_{l'i}\|(\varepsilon_A\|v_{l'j}\| + \varepsilon_c)
\end{equation}
holds, where
\begin{equation}
    \bar B^-_{l'} = n_{l'i}^\top \bar B_l-\left[\textbf{sgn}(u_{l'j}^1)\varepsilon_B\|n_{l'i}\|\ \cdots\ \textbf{sgn}(u_{l'j}^m)\varepsilon_B\|n_{l'i}\|\right ],
\end{equation}
we can conclude that
\begin{equation}
    \begin{aligned}
        \bar B_{l'}u_{l'j} &\ge \bar B_{l'}^-u_{l'j}\\
        &> -n_{l'i}^\top \bar A_{l} v_{l'j} - n_{l'i}^\top \bar c_{l} + \|n_{l'i}\|(\varepsilon_A\|v_{l'j}\| + \varepsilon_c)\\
        &\ge n_{l' i}^\top (\bar A_{l'} v_{l' j} + \bar c_{l'}) 
    \end{aligned}
\end{equation}
holds. By rearranging the terms, the first inequality in \eqref{eq:narrow-ineq-new} holds. We analogously derive the case of $i\in I_-$ by simply enlarging the left-hand side and reducing the right-hand side of the inequality $n_{l'i}^\top \bar B_{l'}u_{l'j} \leq n_{l'i}^\top(\bar A_{l'} v_{l'j} + \bar c_{l'})$. The details are omitted for simplicity.

Thus, we can conclude that for any given $u_{l'j}$, if it lies within the feasible range associated with vertex $v_{l'j}$ specified in the inequality set \eqref{eq:narrow-ineq-new}, then it will also lie within the feasible range defined by inequality set \eqref{eq: simple condition}.

\section*{Appendix C \\ Time Bound Analysis}
For completeness, we restate the derivation of the time bound from \cite{habets2004control}, with adaptation to our setting of a time bound for tiles with unknown but bounded affine dynamics.

Given a set of inputs $u_1, \ldots, u_M$ satisfying the conditions~\eqref{eq:facet-conditions}, an upper bound for the time $T_0$ at which the closed-loop system reaches the exit facet $F_1$ can be obtained. Let $\alpha = \inf\{n_1^\top x|x \in P_N\} = \min\{n_1^\top v_j | j = 1, \ldots, M\}$, and $\beta = \sup\{n_1^\top x|x \in P_N\} = \max\{n_1^\top v_j | j = 1, \ldots, M\}$. Then the function $y(t) = n_1^\top x(t)$ satisfies $\alpha \leq y(0) \leq \beta$ and $y(T_0) = \beta$ because $y(T_0) \in F_1$. Furthermore, $y(t)$ has the following lower bound on its time derivative:
\begin{equation}
    \begin{aligned}
        \inf &\{n_1^\top \dot x(t) | t \in [0, T_0]\} \\
        & \ge \min \{n_1^\top (Ax + B u(x) + c)|x \in P_N\}\\
        & = \min \{\sum_{j = 1}^M \xi_j(x) n_1^\top (A v_j + B u_j + c)|x \in P_N\}\\
        & = \min_{j = 1, \ldots, M} \{n_1^\top (Av_j + B u_j + c)\} = c_1,
    \end{aligned}
\end{equation}
where $\xi_j(x)$ satisfies $\sum_j \xi_j(x) = 1, \xi_j(x)\in [0, 1]$. Thus, $y(t)-y(0) = \int_0^t\dot y(s)ds \ge \int_0^t c_1 ds = c_1t$, $y(t) = y(0)+c_1t$. Let $t=T_0$, $\beta \ge y(0) + c_1T_0$. Since $y(0) \ge \alpha$, we have $\beta \ge \alpha + c_1 T_0$. Then we can obtain the upper bound of $T_0$ as $T_0 \leq (\beta - \alpha)/c_1$.

For another tile $P_{N'}$ with unknown dynamics $A'$, $B'$, $c'$ but satisfies $\|A-A'\|\leq \varepsilon_A$, $\|B-B'\|\leq \varepsilon_B$, and $\|c-c'\|\leq \varepsilon_c$. We seek to find a conservative upper time bound under all admissible $A',B',c'$. Define $\alpha' = \min\{n_1'^\top v_j' | j = 1, \ldots, M\}$, $\beta' = \max\{n_1'^\top v_j' | j = 1, \ldots, M\}$, where $n_1'$ is the normal vector of the expected exit facet of $P_{N'}$ and $v_j'$ denotes the vertices of $P_{N'}$. Then $c_1$ is replaced by $c_1^{\rm rob}$ such that $c_1^{\rm rob} \leq c_1$, which is a more conservative lower bound on the time derivative of $y(t)$. Consider any vertex $v_j$ and corresponding control $u_j$,
\begin{equation}
    \scalebox{0.89}{$\begin{aligned}
    &n_1^\top (A'v_j + B' u_j + c') \\&= n_1^\top [(A v_j + Bu_j + c) 
    + n_1^\top ((A' - A)v_j + (B' - B) u_j + (c' - c)]\\
    &\ge n_1^\top (A v_j + Bu_j + c) - \|n_1\|(\varepsilon_A \|v_j\| + \varepsilon_B\|u_j\| + \varepsilon_c).
\end{aligned}$}
\end{equation}
Taking the minimum of $n_1^\top (A'v_j + B' u_j + c')$ over $j$ yields the worst-case lower bound of $\dot y$ for any given $u_j$ bounded by~\ref{eq:narrow-ineq-new}. Then $c_1^{\rm rob}$ can be chosen to maximize this lower bound over $u_j$:
\begin{equation}
\begin{aligned}
        c_1^{\rm rob} =\max_{u_j}\min_j &[n_1^\top (Av_j + Bu_j + c) \\&- \|n_1\|(\varepsilon_A \|v_j\| + \varepsilon_B\|u_j\|+\varepsilon_c)].
\end{aligned}
\end{equation}
This optimization problem can be solved by an convex optimization solver~\cite{6669541}. Following similar derivation of the upper bound of $T_0$, we have the upper bound of $T_0'$ for $P_{N'}$ as $T_0' \leq (\beta' - \alpha')/c_1^{\rm rob}$.

\section*{Appendix D \\ Formulation of CLF-CBF controller for Target Node}
Once the agent enters the target polytope $P_{\rm target}$, a feedback controller is synthesized to drive the system state to the desired target state $x_{\rm target}$ while ensuring safety with respect to the polytope boundaries. We adopt a Control Lyapunov Function --- Control Barrier Function (CLF-CBF) based controller \cite{7782377, 8796030} for the affine system $\dot x = A_{\rm target}x + B_{\rm target}u + c_{\rm target}$.

To ensure convergence to the target state $x_{\rm target}$, we define the quadratic Lyapunov function
\begin{equation}
    V(x) = \frac{1}{2}(x - x_{\rm target})^\top P (x - x_{\rm target}),
\end{equation}
where $P \in \mathbb{R}^{n \times n}$ is symmetric positive definite. The CLF condition is thus
\begin{equation}
    \dot V(x) = \nabla V(x)^\top (A_{\rm target}x + B_{\rm target}u + c_{\rm target}) \leq -\alpha V(x)
\end{equation}
for some $\alpha >0$, which guarantees asymptotic stability of the equilibrium $x_{\rm target}$.

Safety within the target polytope is enforced through a set of barrier functions associated with its facets. Suppose
\begin{equation}
    P_{\rm target} = \{x \in \mathbb{R}^n | h_i(x) \ge 0, i = 1, \ldots K\},
\end{equation}
where $h_i(x)$ are affine functions that represent the facets. The CBF condition for each facet is given by 
\begin{equation}
    \dot h_i(x) = \nabla h_i(x)^\top (A_{\rm target}x + B_{\rm target}u + c_{\rm target}) \ge -\kappa h_i(x),
\end{equation}
where $\kappa>0$.

The control input is computed by solving the following quadratic program at each time instant:

\begin{equation}
    \scalebox{0.93}{$\begin{aligned} 
           \min_{u \in \mathbb{R}^m} \quad 
& \|u\|^2 + \delta^2 \\
\text{s.t.} \quad 
& \nabla V(x)^\top (A_{\rm target}x + B_{\rm target}u + c_{\rm target}) \leq -\alpha V(x) + \delta, \\
& \nabla h_i(x)^\top (A_{\rm target}x + B_{\rm target}u + c_{\rm target}) \ge -\kappa h_i(x), \\
& \delta \ge 0\\
& u \in P_u,
    \end{aligned}$}
\end{equation}
where $\delta$ is a slack variable that guarantees the feasibility of the program. Naturally, this program does not have a guarantee to stabilize the system to $x_{\rm target}$ since $\delta$ is not necessarily zero.

\bibliographystyle{IEEEtran}
\bibliography{ref}



\end{document}